\documentclass[Afour,sageh,times,doublespace]{sagej}

\usepackage{moreverb,url}

\usepackage{hhline}
\usepackage{multirow}
\usepackage[colorlinks,bookmarksopen,bookmarksnumbered,citecolor=red,urlcolor=red]{hyperref}

\usepackage{algorithm,algcompatible}
\algnewcommand\INPUT{\item[\textbf{Input:}]}%
\algnewcommand\OUTPUT{\item[\textbf{Output:}]}%

\newcommand\BibTeX{{\rmfamily B\kern-.05em \textsc{i\kern-.025em b}\kern-.08em
T\kern-.1667em\lower.7ex\hbox{E}\kern-.125emX}}

\theoremstyle{plain}
\newtheorem{theorem}{Theorem}[]

\theoremstyle{definition}

\DeclareMathOperator*{\argmax}{argmax}
\DeclareMathOperator*{\argmin}{argmin}

\def\volumeyear{2016}

\begin{document}

\runninghead{Liu \textit{et~al.}}

\title{HierMUD: Hierarchical Multi-task Unsupervised Domain Adaptation between Bridges for Drive-by Damage Diagnosis}

\author{Jingxiao Liu\affilnum{1}, Susu Xu\affilnum{2}, Mario Berg\'{e}s\affilnum{3}, and Hae Young Noh\affilnum{1}}

\affiliation{\affilnum{1}Stanford University, CA, USA\\
\affilnum{2}Stony Brook University, NY, USA\\
\affilnum{3}Carnegie Mellon University, PA, USA}

\corrauth{Jingxiao Liu, Department of Civil \& Environmental Engineering,
Stanford University,
Stanford, CA,
94305, USA.}

\email{liujx@stanford.edu}

\begin{abstract}
Monitoring bridges through vibration responses of drive-by vehicles enables efficient and low-cost bridge maintenance by allowing each vehicle to inspect multiple bridges and eliminating the needs for installing and maintaining sensors on every bridge. However, many of the existing drive-by monitoring approaches are based on supervised learning models that require massive labeled data from every bridge. It is expensive and time-consuming, if not impossible, to obtain these labeled data. Furthermore, directly applying a supervised learning model trained on one bridge to new bridges would result in low accuracy due to the shift between different bridges’ data distributions. Moreover, when we have multiple tasks (e.g., damage detection, localization, and quantification), the distribution shifts become more challenging than having only one task because different tasks have distinct distribution shifts and varying task difficulties. 

To this end, we introduce \textbf{HierMUD}, the first \textbf{Hier}archical \textbf{M}ulti-task \textbf{U}nsupervised \textbf{D}omain adaptation framework that transfers the damage diagnosis model learned from one bridge to a new bridge without requiring any labels from the new bridge in any of the tasks. Specifically, our framework learns a hierarchical neural network model in an adversarial way to extract features that are informative to multiple diagnostic tasks and invariant across multiple bridges. To match distributions over multiple tasks, we design a new loss function based on a new provable generalization risk bound to adaptively assign higher weights to tasks with more shifted distributions. 
To learn multiple tasks with varying task difficulties, we split them into easy-to-learn and hard-to-learn based on empirical observations. Then, we formulate a feature hierarchy to utilize more learning resources to improve the hard-to-learn tasks' performance. 
We evaluate our framework with experimental data from 2 bridges and 3 vehicles. We achieve average accuracy of 95\% for damage detection, 93\% for localization, and up to 72\% for quantification.

\end{abstract}


\maketitle

\section{Introduction}
Bridges are key components of transportation infrastructure. Aging bridges all over the world pose challenges to the economy and public safety. According to the 2016 National Bridge Inventory of the Federal Highway Administration, 139,466 of 614,387 bridges in the U.S. are structurally deficient or functionally obsolete \citep{asce}. The state of aging bridges demands researchers to develop efficient and scalable approaches for monitoring a large stock of bridges.

Currently, bridge maintenance is based on manual inspection \citep{hartle2002bridge}, which is inefficient, incurs high labor costs, and fails to detect damages in a timely manner. To address these challenges, structural health monitoring techniques \citep{sun2020review}, where structures are instrumented using sensors (e.g., strain gauge, accelerometers, cameras, etc.) to collect structural performance data, have been developed to achieve continuous and autonomous bridge health monitoring (BHM). Yet, such sensing methods are hard to scale up as they require on-site installation and maintenance of sensors on every bridge and cause interruptions to regular traffic for running tests and maintaining instruments \citep{yang2020state}. 

To address the drawbacks of current BHM, drive-by BHM approaches were proposed to use vibration data of a vehicle passing over the bridge for diagnosing bridge damage. Drive-by BHM is also referred to as the vehicle scanning method \citep{yang2020vehicle} or indirect structural health monitoring \citep{cerda2014indirect, lederman2014damage, liu2020diagnosis}. Vehicle vibrations contain information about the vehicle-bridge interaction (VBI) and thus can indirectly inform us of the dynamic characteristics of the bridge for damage diagnosis~\citep{yang2004extracting,liu2020diagnosis}. This is a scalable sensing approach with low-cost and low-maintenance requirements because each instrumented vehicle can efficiently monitor multiple bridges. Also, there is no need for direct installations and on-site maintenance of sensors on every bridge. 

Previous drive-by BHM focuses on estimating bridge modal parameters (e.g., fundamental frequencies \citep{yang2004extracting,lin2005use,liu2019expectation}, mode shapes \citep{yang2014constructing,malekjafarian2014identification}, and damping coefficients \citep{gonzalez2012identification,liu2019expectation}) that can be used for detecting and localizing bridge damage. However, since the VBI system is a complex physical system involving many types of noise and uncertainties (e.g., environmental noise, vehicle operational uncertainties, etc.), such modal analysis methods for drive-by BHM are susceptible to them \citep{liu2020diagnosis}. This makes the modal parameters' estimation inaccurate and limits the ability of drive-by BHM to diagnose bridge damage (e.g., localize and quantify damage severity). 

More recently, data-driven approaches use signal processing and machine learning techniques to extract informative features from the vehicle acceleration signals~\citep{nguyen2010multi,mcgetrick2013parametric,liu2019expectation,liu2020scalable,eshkevari2020signal,cerda2014indirect,lederman2014damage,liu2020diagnosis,liu2020damage,mei2019indirect,sadeghi2020modal}. The extracted features are more robust to noise, enabling more sophisticated diagnoses such as damage localization and quantification. However, such data-driven approaches generally use supervised learning models developed on available labeled data (i.e., a set of bridges with known damage labels) that is expensive, if not impossible, to obtain. This labeled data requirement is further exacerbated by having multiple diagnostic tasks (e.g., damage detection, localization, and quantification). Furthermore, the standard supervised learning-based approaches learned using vehicle vibration data collected from one bridge are inaccurate for monitoring other bridges because data distributions of the vehicle passing over different bridges are shifted. Having to re-train for each new bridge in multiple bridge monitoring is time-consuming and costly. Therefore, for multiple bridge monitoring with multiple diagnostic tasks, one needs to transfer or generalize the multi-task damage diagnostic model learned from one bridge to other bridges, in order to eliminate the need for requiring training labeled data from every bridge in multiple tasks.

To this end, we introduce \textbf{HierMUD}, a new \textbf{Hier}archical \textbf{M}ulti-task \textbf{U}nsupervised \textbf{D}omain adaptation framework that transfers a model learned from one bridge data to predict multiple diagnostic tasks (including damage detection, localization, and quantification) in another bridge in an unsupervised way. Specifically, HierMUD makes a prediction for the target bridge, without any labels from the target bridge in any of the tasks. We achieve this goal by extracting features that are 1) informative to multiple tasks (i.e., task-informative) and 2) invariant across the source and target domains (i.e., domain-invariant). 

Our framework is inspired by an unsupervised domain adaptation (UDA) approach that has been developed in the machine learning community to address the data distribution shift between two different domains, namely source and target domains~\citep{zhang2019transfer,jiang2007instance,pan2010domain, cao2018unsupervised,luo2020unsupervised,xu2021phymdan}. In our work, we denote the bridge with labeled data as ``Source Domain,” and the new bridge of interest without any labels as ``Target Domain.” UDA focuses on the unsupervised learning tasks in the target domain, which transfers the model learned using source domain data and labels (e.g., vehicle vibration data with the corresponding bridge damage labels) to predict tasks in the target domain without labels (e.g., vehicle vibration data from other bridges without knowing damage labels). In particular, domain adversarial learning algorithm extracts a feature that simultaneously maximizes the damage diagnosis performance for the source domain based on source domain labeled data while minimizing the performance of classifying which domain the feature came from using both source and target domain data~\citep{saito2018maximum,ganin2016domain,zhang2019bridging,xu2021phymdan}. This algorithm can better match complex data distributions between the source and target domains by learning feature representations through neural networks, when compared to other conventional UDA approaches.

However, directly matching different domains' data distributions over each diagnostic task separately through domain adversarial learning limits the overall performance of drive-by BHM, which has multiple diagnostic tasks. This is because these tasks are coupled/related with each other and share damage-sensitive information. The prediction performance of each task depends on that of other coupled tasks. 

Therefore, our algorithm integrates multi-task learning (MTL) with domain-adversarial learning to fuse information from different bridges and multiple diagnostic tasks for effectively improving diagnostic accuracy and scalability of drive-by BHM. MTL is helpful because it simultaneously solves multiple learning tasks to improve the generalization performance of all the tasks, when compared to training the models for each task independently or sequentially (i.e., independent or sequential task learning)~\citep{caruana1997multitask,luong2015multi,augenstein2018multi,dong2015multi,hashimoto2016joint,yang2016deep,misra2016cross,jou2016deep,wan2019bayesian,liu2019damage}.

Yet, when combining MTL and domain-adversarial learning, two particular research challenges exist:
\begin{itemize}
    \item[1)] {\bf Distinct distribution shift:} When we have multiple damage diagnostic tasks, the distribution shift problem becomes more challenging than having only one task because different tasks have distinct shifted distributions between the source and target domains. 
    Therefore, it is important to develop efficient optimization strategies to find an optimal trade-off for matching different domains' distributions over multiple tasks.
    \item[2)] {\bf Varying task difficulty:} The learning difficulties of different tasks vary with the complexity of mappings from the vehicle vibration data distribution to the damage label distributions. Some tasks (e.g., damage quantification task) would have highly non-linear mappings between data and label distributions and, therefore, be more difficult to learn than other tasks (i.e., hard-to-learn). Therefore, we need to distribute learning resources (e.g., representation capacities) according to the learning difficulty of tasks.
\end{itemize}

To address the distinct distribution shift challenge, we introduce a new loss function that prioritizes domain adaptations for tasks having more severe distribution shifts between different domains through a soft-max objective function. To formulate this new loss function, we first derive a new generalization risk bound for multi-task UDA, by optimizing which the new loss function is designed. Specifically, we minimize our loss function to jointly optimize three components: 1) feature extractors, 2) task predictors, and 3) domain classifiers. The parameters of task predictors are optimized to predict task labels in the source domain training set, which ensures that the extracted features are task informative. The parameters of feature extractors are optimized with the domain classifiers in an adversarial way, such that the best trained domain classifier cannot distinguish which domains the extracted features come from. During the optimization, our new loss function adaptively weighs more on minimizing the distribution divergence of tasks having more shifted distributions between different domains. In this way, the model is optimized to automatically find a trade-off for matching different domains' distributions over multiple tasks.

Further, to address the varying task difficulty challenge, we develop hierarchical feature extractors to allocate more learning resources to hard-to-learn tasks. We first split the multiple tasks into easy-to-learn tasks (e.g., damage detection and localization) and hard-to-learn tasks (e.g., damage quantification) based on their degrees of learning difficulty. Specifically, we model the learning difficulty of each task to be inversely proportional to performance in the source domain (e.g., damage localization and quantification accuracy in supervised learning settings). Then, the hierarchical feature extractors learn two-level features: task-shared and task-specific features. To achieve high prediction accuracy for multiple tasks without creating a complex model, we extract task-shared features from input data for easy-to-learn tasks and then extract task-specific features from the task-shared features for only the hard-to-learn tasks. In this way, we allocate more learning resources (learning deeper feature representations) to learn hard-to-learn tasks, which improves the overall performance for all the tasks.

We evaluate our framework on the drive-by BHM application using lab-scale experiments with two structurally different bridges and three vehicles of different weights. In the evaluation, we train our framework using labeled data collected from a vehicle passing over one bridge to diagnose damage in another bridge with unlabeled vehicle vibration data. Our framework outperform five baselines without UDA, MTL, the new loss function, or hierarchical structure.

In summary, this paper has three main contributions:
\begin{itemize}
    \item[1)] We introduce HierMUD, a new multi-task UDA framework that transfers the model learned from one bridge to achieve multiple damage diagnostic tasks in another bridge without any labels from the target bridge in any of the tasks. To the best of our knowledge, this new framework is the first domain adaptation framework for multi-task bridge monitoring. We have since released a PyTorch \citep{paszke2017automatic} implementation of HierMUD at \url{https://github.com/jingxiaoliu/HierMUD}.
    \item[2)] We derive a generalization risk bound that provides a theoretical guarantee to achieve domain adaptation on multiple learning tasks. This work bridges the gaps between the theories and algorithms for multi-task UDA. Based on this bound, we design a new loss function to find a trade-off for matching different domains' distributions over multiple tasks, which addresses the distinct distribution shift challenge. 
    \item[3)] We develop a hierarchical architecture for our multi-task and domain-adversarial learning algorithm. This hierarchical architecture ensures that the framework accurately and efficiently transfers the model for predicting multiple tasks in the target domain, which addresses the varying task difficulty challenge.
\end{itemize}

The remainder of this paper is divided into six sections. In section 2, we study the MTL and data distribution shift challenges in the drive-by BHM application. Section 3 derives the generalization risk bound for multi-task UDA. Section 4 presents our HierMUD framework, which includes the description of our framework, loss function, and algorithm design. Section 5 describes the evaluation of our framework on the drive-by BHM application, following by Section 6 that presents the evaluation results. Section 7 concludes our work and provides discussions about future work.

\section{Data distribution shift and multi-task learning challenges for the drive-by bridge health monitoring}

In this section, we describe the physical insights that enable our drive-by BHM and explain the associated challenges in achieving this scalable BHM approach. We first characterize the structural dynamics of the VBI system. Next, we study the multi-task learning and data distribution shift challenges, respectively, by proving the error propagation between multiple damage diagnostic tasks and characterizing the shifting of the joint distribution of vehicle vibration and damage labels.

\subsection{Characterizing the Structural Dynamics of the VBI System}

\begin{figure*}[htb]
    \centering
    \includegraphics[width=0.8\linewidth]{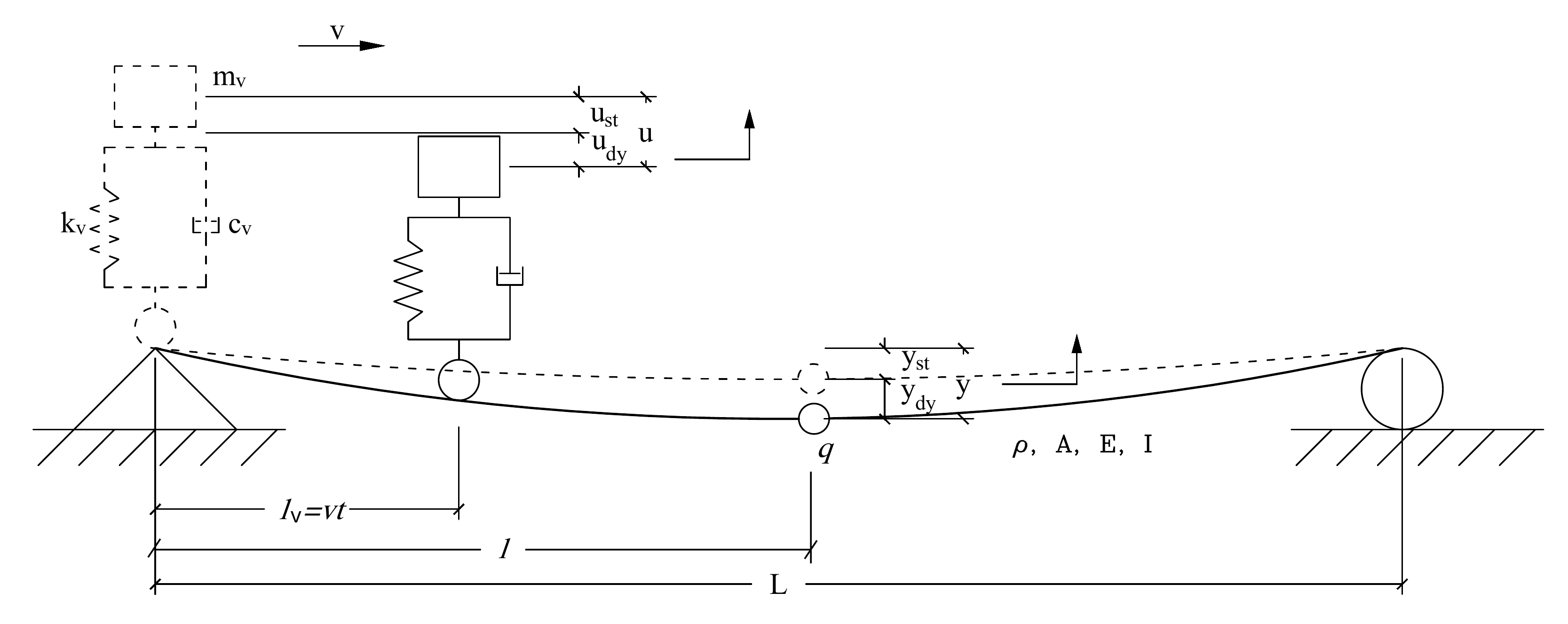}
    \caption{The vehicle-bridge interaction system with a surrogate damage simulated by attaching a mass having magnitude $q$ at location $l$.}
    \label{fig:vbi_model}
\end{figure*}

To provide physical insights of drive-by BHM, we model the VBI system, as shown in Figure~\ref{fig:vbi_model}, as a sprung mass (representing the vehicle) traveling with a constant speed on a simply supported beam (representing the bridge). We assume the beam is of the Euler-Bernoulli type with a constant cross section. We also assume that there is no friction force between the `wheel' and the beam.  The damage is simulated by attaching a mass (magnitude/severity level: $q$) at location ($l$) on the beam. The added mass changes the mass of the bridge and its dynamic characteristics~\citep{malekjafarian2015review}. Modifying the weight of the attached mass is a non-destructive way of creating physical changes to the VBI system to mimic structural damage~\citep{DERAEMAEKER20181,Taddei2018,doi:10.1177/1475921717699375}.

In our prior works~\citep{liu2019damage,liu2020diagnosis}, we have derived the theoretical formulation of the VBI system in the frequency domain, which is summarized in the following paragraphs. 

The equations of motion for the vehicle and bridge in the time domain are first derived as
\begin{equation}
\label{vehicle}
\begin{aligned}
&m_v\ddot{u}_{dy}(t)+k_v[u_{dy}(t)-y_{dy}(x=vt,t)-y_{st}(x=vt)]\\&+c_v[\dot{u}_{dy}(t)-\dot{y}_{dy}(x=vt,t)-\dot{y}_{st}(x=vt)]=0
\end{aligned}
\end{equation}
\begin{equation}
\label{bridge}
\begin{aligned}
&\rho A \ddot{y}_{dy}(x,t)+q\ddot{y}_{dy}(l,t)\delta(x-l)+EIy_{dy}''''(x,t)\\
&=\big\{-m_vg+k_v[u_{dy}(t)-y_{dy}(x=vt,t)-y_{st}(x=vt)]\\
&+c_v[\dot{u}_{dy}(t)-\dot{y}_{dy}(x=vt,t)-\dot{y}_{st}(x=vt)]\big\}\delta(x-vt)
\end{aligned}
\end{equation}
where $m_v,~k_v,~c_v$, and $u$ are the mass, stiffness, damping coefficient, and total displacement of the vehicle, respectively; $\rho,~A,~E,~I$, and $~y_{st}$ are the density, sectional area, Young's modulus, moment of inertia, and the static displacement of the bridge, respectively; $\delta(x-vt)$ is the Dirac delta function; and $u_{dy}(t)$ and $y_{dy}(t)$ are the dynamic displacements of vehicle and bridge, respectively.

Then the $n$-th mode frequency response of the vehicle's acceleration is
\begin{equation}
\begin{aligned}
\label{eq:VBI}
&\ddot{U}_{dy,n}(\omega)=\ddot{U}_{dy,n}\big(\omega-\frac{2n\pi v}{L}\big)\\
&-\frac{i\pi^4EI-i(\omega-\frac{n\pi v}{L})^2L^3(\rho AL+2q\sin^2{(\frac{n\pi l}{L}}))}{2\pi m_vL^3(\omega-\frac{n\pi v}{L})^2\sin\big(\frac{n\pi}{2}\big)}\\
&\times\ddot{Y}_{dy,n}(\frac{L}{2},\omega-\frac{n\pi v}{L})\\
&-\sqrt{2\pi}g\big[\Delta(\omega-\frac{2n\pi v}{L})-\Delta(\omega)\big]
\end{aligned}
\end{equation}
where $Y_{dy,n}(x,\omega)$ is the $n$-th mode frequency response function (FRF) of the bridge element acceleration at the location $x$;  and $U_{dy,n}(\omega)$ and $\Delta(\omega)$ are the $n$-th mode FRF of the vehicle acceleration and the FRF of the Dirac delta function, respectively.

From Equation~\eqref{eq:VBI}, we obtain the following important physical understandings of the drive-by vehicle vibration:
\begin{itemize}
    \item[1)] {\bf Non-linear property}: Vehicle acceleration ($\ddot{U}_{dy,n}(\omega)$) is a high-dimensional signal that has a complex non-linear relationship with bridge properties ($\rho,~A,~E,~I$) and damage parameters ($q,~l$). Thus, it is difficult to infer damage states for different bridges by directly analyzing the raw vehicle signals. It is important to model features that can represent the non-linearity of the VBI system.
    \item[2)] {\bf Coupled diagnostic tasks}: Different damage locations ($l$) and severity levels ($q$) only vary the term $q\sin^2(\frac{n\pi l}{L})$. Let's define the damage information as $d=q\sin^2(\frac{n\pi l}{L})$, representing structural dynamic characteristic changes due to damage. The damage localization and quantification tasks are coupled with each other through the same damage information $d$, and thus the estimation of them depends on each other's estimation.
\end{itemize}

We incorporate these physical insights of the VBI system to develop our multi-task UDA framework. The non-linear property instructs us to use non-linear models or extract non-linear features from the vehicle vibrations for estimating bridge damage. In this work, we use a neural network-based model to non-linearly extract task-informative features from vehicle vibration data.

Moreover, the coupled tasks property of the VBI system informs us to use or extract the shared damage information (e.g., task-shared features) instead of independently learning multiple diagnostic tasks. This is further discussed in the following subsection.

\subsection{Error Propagation between Multiple Tasks for a VBI System}
The shared information among multiple tasks can be learned simultaneously from multiple tasks or learned sequentially from one task to the next. In this section, we illustrate that simultaneous learning (i.e., MTL) is more accurate than sequential learning through a theoretical study of the VBI system. The study shows that the sequential learning method results in a significant error propagation from the previous task to the next.

For instance, if we localize and quantify the bridge damage sequentially (localize the damage first and then quantify the severity of the damage at the obtained damage location), the estimation of damage severity $q$ is $d/\sin^2(\frac{n\pi \hat{l}}{L})$, where  $\hat{l}$ is the estimated damage location. Then, the propagation of error from the damage location estimation to the severity estimation is
\begin{equation}
\begin{aligned}
\label{eq:error_propa}
\sigma_{q}=\pm\sqrt{\sigma_{d}^2\frac{1}{\sin^4(n\pi \hat{l}/L)}+\sigma^2_l\frac{4n^2\pi^2d^2\cos^2(n\pi \hat{l}/L)}{L^2\sin^6(n\pi \hat{l}/L)}}
\end{aligned}
\end{equation}
where $\sigma_q$, $\sigma_l$, $\sigma_{d}$ are errors of severity, location, and damage information estimations, respectively. $\sin^4(n\pi \hat{l}/L)$ and $\sin^6(n\pi \hat{l}/L)$ in the denominators are smaller than 1, which makes the estimation error of $q$ every large as their values decrease. Especially, when the damage is close to the bridge supports (i.e., $|\hat{l}-L/2|\to L/2$), it leads to $sin(n\pi \hat{l}/L)\to 0$ and $\sigma_q\to \infty$. Thus, damage location estimation error propagates, which results in a very inaccurate estimation of damage severity level.

To this end, we solve multiple tasks simultaneously, which can improve the overall accuracy by minimizing error propagation and learning the shared information (e.g., task-shared feature representations) from the coupled tasks. 

Besides simultaneously learning multiple tasks, a scalable drive-by BHM approach needs to work for multiple domains (i.e., bridges). The following subsection discusses the data distribution shift challenge for drive-by monitoring of multiple bridges.

\subsection{Data Distribution Shift for VBI Systems}

\begin{figure}[htb]
    \centering
    \includegraphics[width=1.\linewidth]{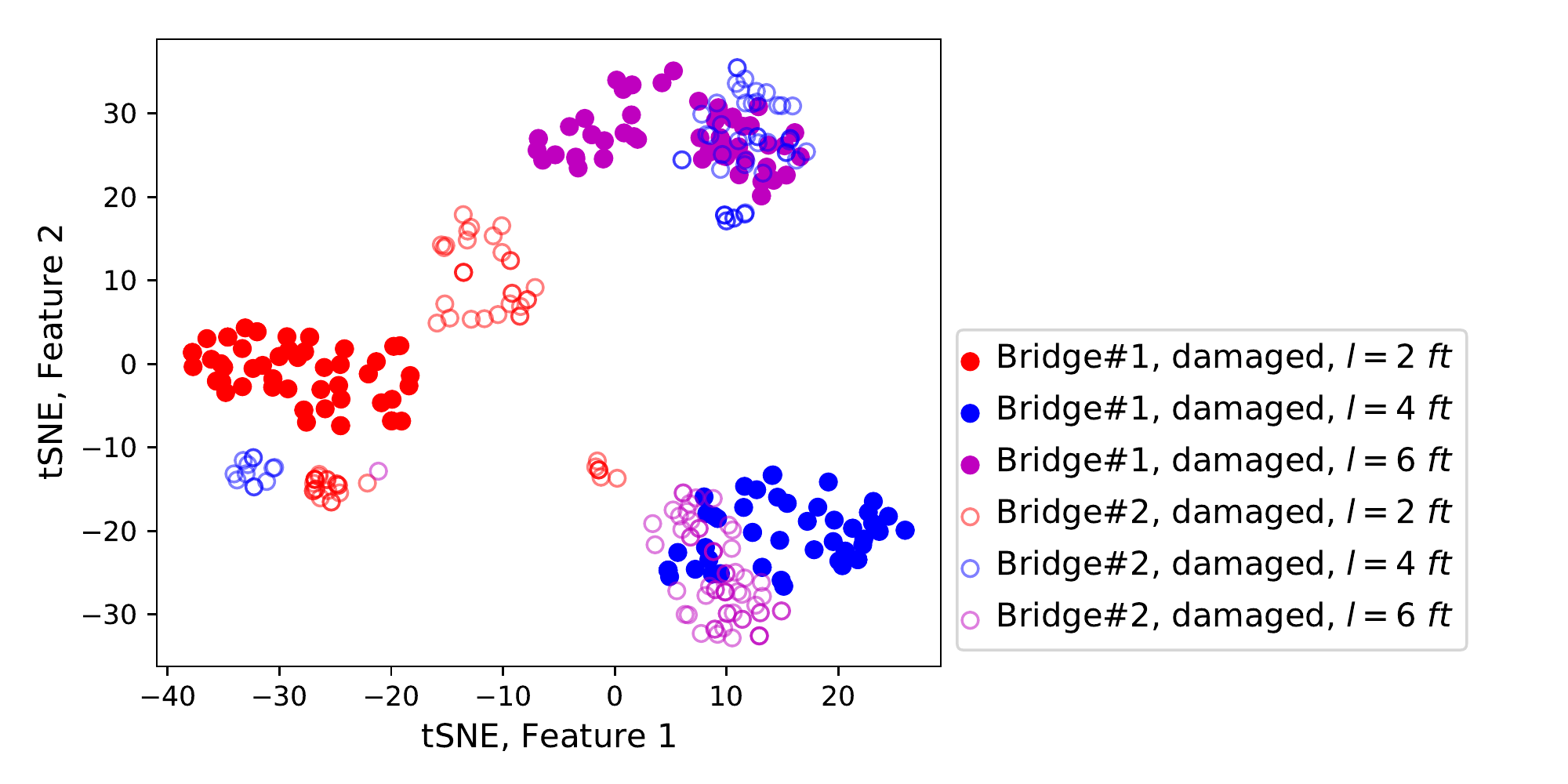}
    \caption{The 2D tSNE visualization of vibration data distributions of a vehicle passing over different bridges. Different colors represent different damage locations on the bridge. Filled markers indicate bridge\#1 data, and unfilled markers indicate bridge\#2 data. This figure shows that directly utilizing the model trained on one bridge's (e.g., Bridge\#1) dataset to predict damage locations of another bridge (e.g., Bridge\#2) can result in very low accuracy due to data distribution shift.}
    \label{fig:beforDA}
\end{figure}

The joint distributions of vehicle vibrations and damage labels are shifted as the vehicle passes by different bridges. If we consider the process of the VBI system as a stochastic process, according to Equation~\ref{eq:VBI}, the joint distributions of the vehicle accelerations ($\ddot{U}_{dy,n}(\omega)$) and damage labels ($q$ or $l$) changes non-linearly with bridge properties (e.g., $\rho,~A,~E,~I$). An example of the shifting of these joint data distributions for different bridges is visualized in a low-dimensional space in Figure~\ref{fig:beforDA}. It shows a two dimensional t-Distributed Stochastic Neighbor Embedding (tSNE)~\citep{van2008visualizing} visualization of vehicle vibration data distributions for Bridges\#1 and \#2. Each vibration signal is collected from a vehicle passing over an 8-feet bridge model with a damage at the location ($l$) of 2 feet, 4 feet, or 6 feet. The data for the two structurally different bridges (Bridge\#1 and \#2) are represented by filled and unfilled markers, respectively. More details of this experiment and dataset are in Evaluation Section. We can observe from Figure~\ref{fig:beforDA} that directly applying the model learned from one bridge's dataset (e.g., Bridge\#1) to localize damage on the other bridge (e.g., Bridge\#2) results in a low prediction accuracy because of the joint distribution shift.

To address the distribution shift challenge and achieve a scalable drive-by BHM that is invariant across multiple bridges (i.e., it can predict damage without requiring training data from every bridge), we introduce a new multi-task UDA approach. In the next section, we first investigate the multi-task UDA problem theoretically through deriving a generalization risk bound.



\section{A Generalization Risk Bound for Multi-task Unsupervised Domain Adaptation}
In this section, we derive the upper bound of the generalization risk for multi-task UDA problems to investigate the theoretical guarantee of its performance on target domain unseen data. The generalization risk (or error) of a model is the difference between the empirical loss on the training set and the expected loss on a test set, as defined in statistical learning theory \citep{jakubovitz2019generalization}. In other words, it represents the ability of the trained model to generalize from the training dataset to a new unseen dataset. In our problem, the generalization risk is defined to represent how accurately a classifier trained using source domain labeled data and target domain unlabeled data predicts class labels in the target domain. Therefore, deriving the upper bound of the generalization risk provides insights on how to develop learning algorithms to efficiently optimize it.

We first derive a generalization risk bound for UDA and then integrate it with the risk bound for MTL. Next, we characterize the newly derived generalization risk bound to provide insights to our multi-task UDA problem.

\subsection{A generalization risk bound for unsupervised domain adaptation}
We first derive a new generalization risk bound for UDA by representing the original data distribution in a feature space, which has been ignored by the existing risk bounds for UDA~\citep{ganin2016domain,zhang2019bridging,zhao2018adversarial}. Having a feature space enables the modeling of task-shared feature representation when we have multiple tasks. This results in a tighter generalization risk bound for multi-task UDA than independently estimating each task's generalization risk bound~\citep{maurer2016benefit}. Yet, this feature space requires us to estimate the discrepancy between the marginal feature distributions of the source and target domains for obtaining the generalization risk bound, which is introduced in this section.

We consider a classification task that labels input $X$ as belonging to different classes $Y$. We also consider mappings
$$X\xrightarrow[]{w} Z\xrightarrow[]{h} Y,$$
where $X$, $Z$, and $Y$ are random variables of input, feature representation, and class label, which are taken from the input, feature, and output space $\mathcal{X}$, $\mathcal{Z}$, and $\mathcal{Y}$, respectively. The function $w:\mathcal{X}\to \mathcal{Z}$ is a $k$-dimension feature transformation, and the function $h:\mathcal{Z}\to \mathcal{Y}$ is a hypothesis on the feature space (i.e., a labeling function). Then, we have a predictor $h\circ w$, that is, $(h\circ w)(x)=h(w(x))$, for every $x\in\mathcal{X}$. 

Further, we define a domain as a distribution on the input space $\mathcal{X}$ and the output space $\mathcal{Y}$. UDA problems involve two domains, a source domain $\mathcal{D}_S$ and a target domain $\mathcal{D}_T$. We denote $\mathcal{D}_S^X$, $\mathcal{D}_T^X$, $\mathcal{D}_S^Y$, and $\mathcal{D}_T^Y$ as the marginal data ($X$) and label ($Y$) distributions in the source ($\mathcal{D}_S$) and target ($\mathcal{D}_T$) domains, respectively. Note that we also have feature representation distributions, $\mathcal{D}_S^Z$ and $\mathcal{D}_T^Z$, in the source and target domains. Mathematically, an unsupervised domain adaptation algorithm has independent and identically distributed (i.i.d.) labeled source samples $<{X}_S,{Y}_S>$ drawn from $\mathcal{D}_S$ and i.i.d. unlabeled target samples ${X}_T$ drawn from $\mathcal{D}_T^X$, as shown below: 
$$<{X}_S,{Y}_S>=\{x_i,y_i\}_{i=1}^{n_S}\sim \mathcal{D_S};$$
$${X}_T=\{x_i\}_{i=1}^{n_T}\sim\mathcal{D}_T^X$$
where $n_S$ and $n_T$ are the number of samples in the source and target domains, respectively. The goal of UDA is to learn $h$ and $w$ with a low target domain risk under distribution $\mathcal{D}_T$, which is defined as: $\epsilon_T(h\circ w,f_T)=\text{Pr}_{x\sim \mathcal{D}_T^X}(h\circ w(x)\neq f_T(x)),$
where $f_T$ is the ground truth labeling function and $y=f_T(x)~\text{for~}(x,y)\sim\mathcal{D}_T$.

Since we do not have labeled data in the target domain, we cannot directly compute the target domain risk. Therefore, the upper bound of the target domain risk is estimated by the source domain risk and the discrepancy between the marginal data distributions of the source and target domains, $\mathcal{D}_S^X$ and $\mathcal{D}_T^X$. The discrepancy between $\mathcal{D}_S^X$ and $\mathcal{D}_T^X$ is quantified through the $\mathcal{H}\Delta\mathcal{H}$-divergence~\citep{ben2010theory} that measures distribution divergence with finite samples of unlabeled data from $\mathcal{D}_S^X$ and $\mathcal{D}_T^X$. It is defined as 
\begin{align*}
    d_{\mathcal{H}\Delta\mathcal{H}}(\mathcal{D}_S^X,\mathcal{D}_T^X)=2\sup_{h,h'\in\mathcal{H}}|&\text{Pr}_{x\sim\mathcal{D}_S^X}[h(x)\neq h'(x)]\\&-\text{Pr}_{x\sim\mathcal{D}_T^X}[h(x)\neq h'(x)]|,
\end{align*}
where $\mathcal{H}$ is a hypothesis space, and $\mathcal{H}\Delta\mathcal{H}$ is the symmetric difference hypothesis.


Then, we can derive the generalization bound for UDA in Theorem~\ref{thm:1}.
\begin{theorem}
Let $\mathcal{W}$ be a hypothesis space on $\mathcal{X}$ with VC dimension $d_W$ and $\mathcal{H}$ be a hypothesis space on $\mathcal{Z}$ with VC dimension $d_H$. If ${X}_S$ and ${X}_T$ are samples of size $N$ from $\mathcal{D}_S^X$ and $\mathcal{D}_T^X$, respectively, and ${Z}_S$ and ${Z}_T$ follow distributions $\mathcal{D}_S^Z$ and $\mathcal{D}_T^Z$, respectively,, then for any $\delta\in(0,1)$ with probability at least $1-\delta$, for every $h\in\mathcal{H}$ and $w\in\mathcal{W}$:
\begin{align*}
    \epsilon_T(h\circ w;f_T)
    \leq& \epsilon_S(h\circ w;f_S)+2\epsilon_S(h\circ w^*,f_S)\\&+\frac{1}{2}d_{\mathcal{H}\Delta\mathcal{H}}({Z}_T,{Z}_S)\\
    &+\mathcal{O}\Big(\sqrt{\frac{2d_H\log{(2N)+\log(2/\delta)}}{N}}\Big)\\
    &+\frac{1}{2}\sup_{\hat{h}\in\mathcal{H}}\big[d_{\hat{h},\mathcal{W}\Delta\mathcal{W}}({X}_S,{X}_T)\big]\\
    &+\mathcal{O}\Big(\sqrt{\frac{2d_W\log{(2N)+\log(2/\delta)}}{N}}\Big)\\
    &+\epsilon_T(h^*\circ w^*;f_T)+\epsilon_S(h^*\circ w^*;f_S),
\end{align*}
where 
$$w^*,h^*=\argmin_{\substack{w^*\in\mathcal{W},h^*\in\mathcal{H}}}\epsilon_{S}(h\circ w;f_S)+\epsilon_{T}(h\circ w;f_T),$$
\begin{align*}
    d_{h,\mathcal{W}\Delta\mathcal{W}}&({\mathcal{D}}_S^X,{\mathcal{D}}_T^X)\\
    =&2\sup_{w,w'\in\mathcal{W}}|\text{Pr}_{x\sim\mathcal{D}_T^X}[h\circ w(x)\neq h\circ w'(x)]\\
    &-\text{Pr}_{x\sim\mathcal{D}_S^X}[h\circ w(x)\neq h\circ w'(x)]|\\
    \leq&d_{h,\mathcal{W}\Delta\mathcal{W}}({X}_S,{X}_T)\\
    &+\mathcal{O}\Big(\sqrt{\frac{2d_W\log{(2N)+\log(2/\delta)}}{N}}\Big).
\end{align*}

\begin{proof}
See Appendix.
\end{proof}
\label{thm:1}
\end{theorem}

We prove in Theorem~\ref{thm:1} that the upper bound of the target domain risk consists of five components: 
\begin{itemize}
    \item[1)] The source domain risk: $$\epsilon_S(h\circ w;f_S)+2\epsilon_S(h\circ w^*,f_S),$$ which quantifies the error for estimating class labels in the source domain.
    \item[2)] The minimal risk: $$\epsilon_T(h^*\circ w^*;f_T)+\epsilon_S(h^*\circ w^*;f_S),$$ which quantifies the error for estimating class labels using the ideal joint hypothesis over the source and target domains. It is the smallest error we can achieve using the best predictor in the hypothesis set. 
    \item[3)] The empirical symmetric divergence between marginal feature distributions: $$\frac{1}{2}d_{\mathcal{H}\Delta\mathcal{H}}({Z}_T,{Z}_S),$$ which quantifies the distribution difference between the source and target domain marginal feature distributions.
    \item[4)] The supremum of empirical symmetric divergence set between marginal data distributions: $$\frac{1}{2}\sup_{\hat{h}\in\mathcal{H}}\big[d_{\hat{h},\mathcal{W}\Delta\mathcal{W}}({X}_S,{X}_T)\big],$$ which quantifies the distribution difference between the source and target domain marginal data distributions.
    \item[5)] The Big-O terms that measure the complexity of the estimation of divergence.
\end{itemize} 

Next, in the following subsection, we derive a generalization risk bound for multi-task UDA problems by considering the feature space $\mathcal{Z}$ being the task-shared feature space for multiple tasks.

\subsection{Integrating multi-task learning bound with the unsupervised domain adaptation bound}
We first consider multiple classification tasks that label input $X_m$ as belonging to different classes $Y_m$, for $m=1,2,\cdots,M$, where $M$ is the total number of tasks. The mappings for this multi-task learning problem becomes
$$X_m\xrightarrow[]{w} Z_m\xrightarrow[]{h_m} Y_m,~\text{for~}m=1,2,\cdots,M,$$
where $X_m$, $Z_m$, and $Y_m$ are the $m$-th task's random variables of input, feature representation, and class label, respectively, which are taken from the input, feature, and output space $\mathcal{X}$, $\mathcal{Z}$, and $\mathcal{Y}$, respectively. The function $w$ is a task-shared k-dimensional feature transformation, and the function $h_m$ is a task-specific hypothesis for the $m$-th task on the feature space. For each task, we have a predictor $h_m\circ w$. We define the task-averaged true risk under the joint distribution $\prod_{m=1}^M\mathcal{D}_m$ as
\begin{align*}
    \epsilon_{avg}(w,h_1,&\cdots,h_M;f_{1},\cdots,f_{M})\\
    &=\frac{1}{M}\sum_{m=1}^M\text{Pr}_{x\sim \mathcal{D}_{m}^X}(h_m\circ w(x)\neq f_{m}(x)),
\end{align*}
where $f_{m}$ is the ground truth labeling function for the $m$-th task and $y=f_{m}(x)$ for $(x,y)\sim\mathcal{D}_m$. 
We also define the task-averaged empirical risk as
\begin{align*}
    \epsilon_{avg}(\hat{w},\hat{h}_1,&\cdots,\hat{h}_M;\bar{Y})\\
    &=\frac{1}{NM}\sum_{m=1}^M\sum_{i=1}^N\mathbb{I}[\hat{h}_m(\hat{w}(x_{m,i}))\neq y_{m,i}],
\end{align*}
where $N$ is the total number of samples in each task, which is assumed to be the same for each task, and
$\mathbb{I}(\cdot)$ is the indicator function. We define that $(\bar{X},\bar{Y})=({X}_{1},\cdots,{X}_{M},{Y}_{1},\cdots,{Y}_{M})$ are i.i.d. samples drawn from the joint distribution $\prod_{m=1}^M\mathcal{D}_{m}$. For the $m$-th task,
$$({X}_m,{Y}_m)=\{x_{m,i},y_{m,i}\}_{i=1}^N\sim\mathcal{D}_m.$$

Further, we consider a multi-task UDA problem, which has labeled samples $(\bar{X}_S,\bar{Y}_S)=({X}_{S,1},\cdots,{X}_{S,M},{Y}_{S,1},\cdots,{Y}_{S,M})$ drawn from a joint source domain $\prod_{m=1}^M\mathcal{D}_{S,m}$ and unlabeled samples $\bar{X}_T=({X}_{T,1},\cdots,{X}_{T,M})$ drawn from a joint target domain $\prod_{m=1}^M\mathcal{D}_{T,m}^X$. The goal of multi-tasks UDA is to learn $h_1,\cdots,h_M$ and $w$ with a low target domain task-averaged risk under the joint distribution $\prod_{m=1}^M\mathcal{D}_{T,m}^X$, which is defined as: $\epsilon_{avg,T}(w,h_{1},\cdots,h_{M};f_{T,1},\cdots,f_{T,M}).$

Our generalization risk bound for multi-task UDA is built on our Theorem~\ref{thm:1} by combining it with the risk bound for multi-task learning. The multi-task learning risk bound was introduced in the work of~\cite{maurer2016benefit}, which showed that the upper bound of the task-averaged risk consists of the task-averaged empirical risk, the complexity measure relevant to the estimation of the representation, and the complexity measure of estimating task-specific predictors.
Specifically, with probability at least $1-\delta$, where $\delta\in (0,1)$, in the draw of $(\bar{X},\bar{Y})=({X}_1,\cdots,{X}_M,{Y}_1,\cdots,{Y}_M)\sim \prod_{m=1}^M\mathcal{D}_m$, it holds for every $w\in \mathcal{W}$ and every $h_1,\cdots,h_M\in\mathcal{H}$ that
\begin{equation}
    \begin{aligned}
\label{eq:MTL_bound}
    \epsilon_{avg}(w,h_1,&\cdots,h_M;f_{1},\cdots,f_{M})\\
    \leq& \epsilon_{avg}(\hat{w},\hat{h}_1,\cdots,\hat{h}_M;\bar{Y})\\
    &+c_1\frac{L\hat{G}(\mathcal{W}(\bar{X}))}{NM}+c_2\frac{Q\sup_{w\in\mathcal{W}}\|w(\bar{X})\|}{N\sqrt{M}}\\
    &+\sqrt{\frac{9\log(2/\delta)}{2NM}},
    \end{aligned}
\end{equation}
where $L$ is the Lipschitz constant for $h\in\mathcal{H}$; $c_1$ and $c_2$ are universal constants; $\hat{G}(\mathcal{W}(\bar{X}))$ is the Gaussian average that measures the empirical complexity relevant to the estimation of the feature representation; and
$Q=\sup_{z\neq z'\in\mathcal{Z}}\frac{1}{\|z-z'\|}\mathbb{E}\sup_{h\in\mathcal{H}}\sum_{i=1}^N\gamma_i(h(z_i)-h(z_i'))$

Now, by integrating the multi-task learning risk bound in Equation~\eqref{eq:MTL_bound} with the unsupervised domain adaptation risk bound in Theorem~\ref{thm:1}, we obtain the following theorem:
\begin{theorem}
Let $\mathcal{D}_{S,1},\cdots \mathcal{D}_{S,M}$ and $\mathcal{D}_{T,1}^X,\cdots \mathcal{D}_{T,M}^X$ be probability measure on $(\mathcal{X},\mathcal{Y})$. Let $\mathcal{W}$ be a hypothesis space on $\mathcal{X}$ with VC dimension $d_W$ and $\mathcal{H}$ be a hypothesis space on $\mathcal{Z}$ with VC dimension $d_H$. Let $\delta\in(0,1)$. With probability at least $1-\delta$ in the draw of $(\bar{X}_S,\bar{Y}_S)\sim \prod_{m=1}^M\mathcal{D}_{S,m}$, $\bar{X}_T\sim \prod_{m=1}^M\mathcal{D}_{T,m}^X$ (i.e., $({X}_{S,m},{Y}_{S,m})\sim \mathcal{D}_{S,m}$ and ${X}_{T,m}\sim \mathcal{D}_{T,m}^X$ for $m=1,\cdots, M$), and $\bar{Z}_S,~\bar{Z}_T$ that follow distributions $\prod_{m=1}^M\mathcal{D}_{S,m}^Z$ and $\prod_{m=1}^M\mathcal{D}_{T,m}^Z$, it holds for every $w\in \mathcal{W}$ and every $h_1,\cdots,h_M\in\mathcal{H}$ that
\begingroup
\allowdisplaybreaks
\begin{equation}
\label{eq:bound}
\begin{aligned}
    \epsilon&_{avg,T}(w,h_{1},\cdots,h_{M};f_{T,1},\cdots,f_{T,M})\\
    \leq & \epsilon_{avg,S}(\hat{w},\hat{h}_{1},\cdots,\hat{h}_{M};\bar{Y}_S)\\
    &+c_1\frac{L\hat{G}(\mathcal{W}(\bar{X}_S))}{NM}+c_2\frac{Q\sup_{w\in\mathcal{W}}\|w(\bar{X}_S)\|}{N\sqrt{M}}\\
    &+2\epsilon_{avg,S}(w^*,\hat{h}_{1},\cdots,\hat{h}_{M};\bar{Y}_S)+\frac{2\sqrt{2\pi}\hat{G}(\mathcal{H}(\bar{Z}_S))}{NM}\\
    &+\frac{1}{2}d_{\mathcal{H}\Delta\mathcal{H}}(\bar{Z}_S,\bar{Z}_T)+\mathcal{O}\Big(\sqrt{\frac{2d_H\log{(2N)+\log(2/\delta)}}{N}}\Big)\\
    &+\sup_{\hat{h}_1\cdots \hat{h}_M\in\mathcal{H}}\Big[\frac{1}{2}d_{\hat{h}_1,\cdots,\hat{h}_M,\mathcal{W}\Delta\mathcal{W}}(\bar{X}_S,\bar{X}_T)\Big]\\
    &+\mathcal{O}\Big(\sqrt{\frac{2d_W\log{(2N)+\log(2/\delta)}}{N}}\Big)\\
    &+\epsilon_{avg,S}(w^*,h_{1}^*,\cdots,h_{M}^*;\bar{Y}_S)\\
    &+\epsilon_{avg,T}(w^*,h_{1}^*,\cdots,h_{M}^*;f_{T,1},\cdots,f_{T,M}),
\end{aligned}
\end{equation}
\endgroup
where 
\begin{align*}
w^*,h_{1}^*,&\cdots,h_{M}^*\\
=&\argmin_{\substack{w^*\in\mathcal{W},h_{1}^*,\cdots,h_{M}^*\in\mathcal{H}}}\big[\epsilon_{avg,S}(w,h_{1},\cdots,h_{M};\bar{Y}_S)\\
&+\epsilon_{avg,T}(w,h_{1},\cdots,h_{M};f_{T,1},\cdots,f_{T,M})\big].
\end{align*}
\label{thm:3}
\end{theorem}
We show in Theorem~\ref{thm:3} that the upper bound of task-averaged target domain risk contains seven components:

\begin{itemize}
    \item[1)] The source domain empirical risks:
\begin{align*}
    &\epsilon_{avg,S}(\hat{w},\hat{h}_{1},\cdots,\hat{h}_{M};\bar{Y}_S)\\&+2\epsilon_{avg,S}(w^*,\hat{h}_{1},\cdots,\hat{h}_{M};\bar{Y}_S),
\end{align*}
\item[2)]
the task-averaged minimal risks:
\begin{align*}
    &\epsilon_{avg,S}(w^*,h_{1}^*,\cdots,h_{M}^*;\bar{Y}_S)\\&+\epsilon_{avg,T}(w^*,h_{1}^*,\cdots,h_{M}^*;f_{T,1},\cdots,f_{T,M}),
\end{align*} 
\item[3)]
the complexity measure relevant to the estimation of the representation: $$c_1\frac{L\hat{G}(\mathcal{W}(\bar{X}_S))}{NM},$$ 
\item[4)] the complexity measure of estimating task-specific predictors: $$c_2\frac{Q\sup_{w\in\mathcal{W}}\|w(\bar{X}_S)\|}{N\sqrt{M}}+\frac{2\sqrt{2\pi}\hat{G}(\mathcal{H}(\bar{Z}_S))}{NM},$$ 
\item[5)]
the empirical symmetric divergence between marginal feature distributions: $$\frac{1}{2}d_{\mathcal{H}\Delta\mathcal{H}}(\bar{Z}_S,\bar{Z}_T),$$ 
\item[6)]
the supremum of empirical symmetric divergence set (for multiple tasks) between marginal data distributions: $$\sup_{\hat{h}_1\cdots \hat{h}_M\in\mathcal{H}}\Big[\frac{1}{2}d_{\hat{h}_1,\cdots,\hat{h}_M,\mathcal{W}\Delta\mathcal{W}}(\bar{X}_S,\bar{X}_T)\Big],$$
\item[7)] the Big-O complexity measures of the estimation of divergence.
\end{itemize}

Once we determine the hypothesis sets $\mathcal{H}$ and $\mathcal{W}$, the task-averaged minimal risk and complexity terms are fixed~\citep{zhao2018adversarial}. Therefore, we can minimize the target domain risk bound (Equation~\ref{eq:bound}) by minimizing the sum of the source domain empirical risks, the empirical divergence between marginal data distributions, and the empirical divergence between marginal feature distributions:
\begin{equation}
\label{eq:loss0}
\begin{aligned}
    \text{minimize}~ \Big(&\epsilon_{avg,S}(\hat{w},\hat{h}_{1},\cdots,\hat{h}_{M};\bar{Y}_S)\\
    &+2\epsilon_{avg,S}(w^*,\hat{h}_{1},\cdots,\hat{h}_{M};\bar{Y}_S)\\
    &+\frac{1}{2}d_{\mathcal{H}\Delta\mathcal{H}}(\bar{Z}_S,\bar{Z}_T)\\
    &+\sup_{\hat{h}_1\cdots \hat{h}_M\in\mathcal{H}}\Big[\frac{1}{2}d_{\hat{h}_1,\cdots,\hat{h}_M,\mathcal{W}\Delta\mathcal{W}}(\bar{X}_S,\bar{X}_T)\Big]\Big)
\end{aligned}
\end{equation}

To efficiently solve Equation~\eqref{eq:loss0} that minimizes the generalization risk bound for multi-task UDA, we interpret and characterize the bound in the following subsection.

\subsection{Characterizing the derived risk bound}
We can learn two main insights of the new generalization risk bound for multi-task UDA from Equation~\eqref{eq:loss0}:
\begin{itemize}
\item[1)] {\bf Feature divergence minimization:} Some UDA methods (e.g.,~\citealt{zhang2019bridging,saito2018maximum}) minimize the empirical symmetric divergences between marginal feature distributions (i.e., $\frac{1}{2}d_{\mathcal{H}\Delta\mathcal{H}}(\bar{Z}_S,\bar{Z}_T)$) to make the task-specific classifiers (i.e., $h_1,\cdots, h_M$) invariant across domains. However, these methods are not scalable as the number of tasks grows because they require every task-specific classifier to be domain-invariant. Therefore, our multi-task UDA approach avoids directly minimizing the feature divergence, which requires adapting classifiers of each task separately.

\item[2)] {\bf Data divergence minimization:} Some UDA methods (e.g.,~\citealt{ganin2016domain,long2015learning}) minimize the empirical symmetric divergence between marginal data distributions (i.e., $\sup_{\hat{h}_1\cdots \hat{h}_M\in\mathcal{H}^M}\Big[\frac{1}{2}d_{\hat{h}_1\cdots,\hat{h}_M,\mathcal{W}\Delta\mathcal{W}}(\bar{X}_S,\bar{X}_T)\Big]$) to extract domain-invariant features. Such methods would be successful and scalable if the feature distributions $\mathcal{D}_S^Z$ and $\mathcal{D}_T^Z$ are matched because in this case the empirical symmetric divergence between marginal feature distributions (i.e., $\frac{1}{2}d_{\mathcal{H}\Delta\mathcal{H}}(\bar{Z}_S,\bar{Z}_T)$) would be also very small or even be zero. However, directly minimizing this supremum divergence is difficult and data inefficient under the MTL setting because different tasks have distinct distribution shifts between the source and target domains. Therefore, we introduce a new efficient optimization strategy to find an optimal trade-off for minimizing the data divergence over multiple tasks.
\end{itemize}


In summary, to develop an algorithm that is scalable to the number of tasks, we need to minimize the empirical divergence between marginal data distributions to learn a feature mapping that matches feature distributions between different domains. Therefore, we can rewrite Equation~\eqref{eq:loss0} as
\begin{equation}
\label{eq:loss1}
\begin{aligned}
    \text{minimize}~ \Big(&\epsilon_{avg,S}(\hat{w},\hat{h}_{1},\cdots,\hat{h}_{M};\bar{Y}_S)\\
    &+2\epsilon_{avg,S}(w^*,\hat{h}_{1},\cdots,\hat{h}_{M};\bar{Y}_S)\\
    &+\sup_{\hat{h}_1\cdots \hat{h}_M\in\mathcal{H}}\Big[\frac{1}{2}d_{\hat{h}_1,\cdots,\hat{h}_M,\mathcal{W}\Delta\mathcal{W}}(\bar{X}_S,\bar{X}_T)\Big]\Big)
\end{aligned}
\end{equation}

For a classification problem, minimizing the first two terms in Equation~\eqref{eq:loss1} can be achieved by minimizing the cross-entropy loss between predicted labels and ground truth labels in source domain: $\hat{\epsilon}_{avg,S}=$
$$-\frac{1}{M}\sum_{m=1}^M\mathbb{E}_{(x,y)\sim (X_{S,m},Y_{S,m})}\sum_{c=1}^{C_{m}}\mathbb{I}(y=c)\log \hat{h}_m(\hat{w}(x)),$$
where $C_m$ is the number of classes for the $m$-th tasks.

Further, if we consider $\hat{h}_1,\cdots,\hat{h}_M$ are hypotheses independently drawn from the hypothesis class $\mathcal{H}$, we can write the last term in Equation~\eqref{eq:loss1} as 
\begin{equation}
\label{eq:divergence}
    \max_{m\in [M]}\sup_{\hat{h}_m\in\mathcal{H}}\Big[\frac{1}{2}d_{\hat{h}_m,\mathcal{W}\Delta\mathcal{W}}(X_{S,m},X_{T,m})\Big],
\end{equation} without loss of generality. This means that we can minimize the maximum divergence between marginal feature distributions over all the tasks to achieve the minimization of the divergence term in Equation~\eqref{eq:loss1}. An approximation of the empirical symmetric divergence between distributions is computed by learning a domain discriminator ($h_m\circ w$) that distinguishes samples from different domains ~\citep{ganin2016domain}:
\begin{equation}
\label{eq:divergence1}
\begin{aligned}
    d&_{h_m,\mathcal{W}\Delta\mathcal{W}}({X}_{S,m},{X}_{T,m})\\
    =&2\Big(1-\min_{w\in\mathcal{W}\Delta\mathcal{W}}\big(\frac{1}{N}\sum_{x\sim X_{S,m}}\mathbb{I}(h_m(w(x)=1)\\
    &+\frac{1}{N}\sum_{x\sim X_{T,m}}\mathbb{I}(h_m(w(x)=0)\big)\Big).
\end{aligned}
\end{equation}

To this end, there are three ways to minimize Equation~\eqref{eq:loss1}: 
\begin{itemize}
    \item[1)] \emph{Hard-max objective}: directly minimizing the maximum divergence over all $M$ tasks:
    \begin{equation}
    \label{eq:loss2}
    \begin{aligned}
        &\text{minimize}~ \Big(\hat{\epsilon}_{avg,S}\\
        &+\max_{m\in [M]}\sup_{\hat{h}_m\in\mathcal{H}}\Big[\frac{1}{2}d_{\hat{h}_m,\mathcal{W}\Delta\mathcal{W}}(X_{S,m},X_{T,m})\Big]\Big);
    \end{aligned}
    \end{equation}
    \item[2)] \emph{Average objective}: minimizing the average divergence:
    \begin{equation}
    \label{eq:loss3}
    \begin{aligned}
        &\text{minimize}~ \Big(\hat{\epsilon}_{avg,S}\\
        &+\frac{1}{2M}\sum_{m=1}^M \sup_{\hat{h}_m\in\mathcal{H}}\Big[d_{\hat{h}_m,\mathcal{W}\Delta\mathcal{W}}(X_{S,m},X_{T,m})\Big]\Big);
    \end{aligned}
    \end{equation}
    \item[3)] \emph{Soft-max objective}: minimizing a soft maximum of Equation~\eqref{eq:divergence1}:
    \begin{equation}
    \label{eq:loss4}
    \begin{aligned}
        \text{minimize}~ \Big(&\hat{\epsilon}_{avg,S}+\frac{1}{2}\log\sum_{m=1}^M\exp(\\
        &\sup_{\hat{h}_m\in\mathcal{H}}\Big[d_{\hat{h}_m,\mathcal{W}\Delta\mathcal{W}}(X_{S,m},X_{T,m})\Big])\Big).
    \end{aligned}
    \end{equation}
\end{itemize}
The hard-max objective is data inefficient because the gradient of the max function is only non-zero for $h_m$ with the maximum divergence, and the algorithm only updates its parameters based on the gradient from one of the $M$ tasks. The average objective updates algorithm parameters based on the average gradient from all $M$ tasks. However, this objective considers each task as equally contributing to updating the algorithm parameters, which may not allocate enough computational and learning resources to optimize tasks with larger divergence. The soft-max objective combines the gradients from all the tasks and adaptively assigns the loss of task that have a larger divergence with a heavier weight \citep{zhao2018adversarial}. In this way, the model automatically applies larger gradient magnitudes to tasks having more shifted distributions between different domains. As a result, we propose to use the soft-max objective (Equation~\ref{eq:loss4}) to optimize the multi-task UDA problem. 

\section{HierMUD: A Hierarchical Multi-Task Unsupervised Domain Adaptation Framework}

\begin{figure*}[htb]
    \centering
    \includegraphics[width=0.9\linewidth]{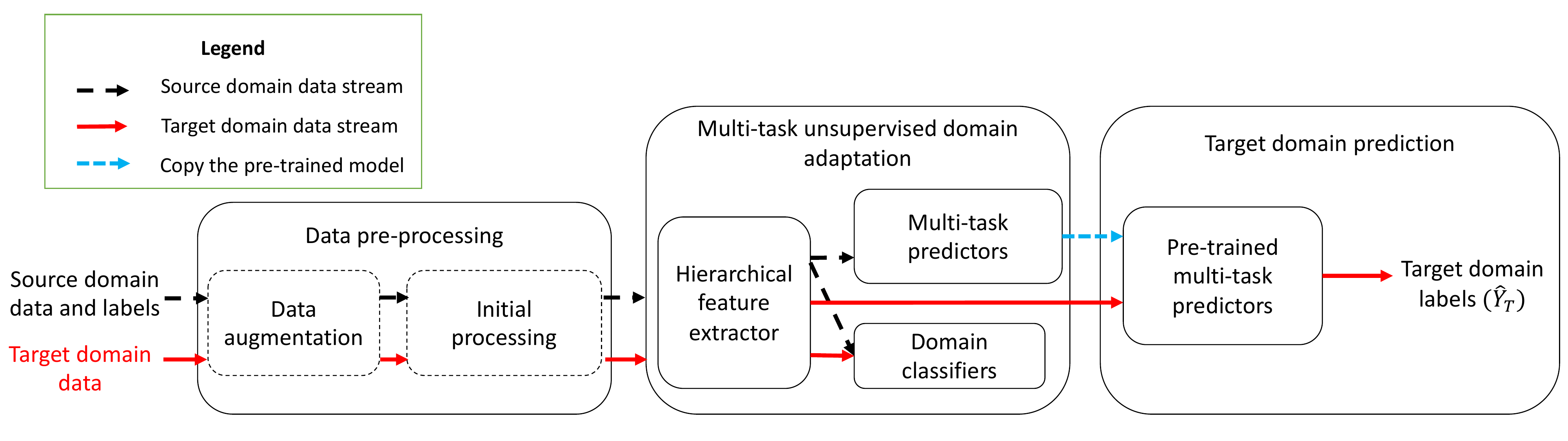}
    \caption{The flowchart for our HierMUD framework. The input to the framework is source domain data with the corresponding labels and the unlabeled target domain data, and the framework outputs the predicted target domain labels. Black dash lines indicate source domain data stream, solid red lines indicate target domain data stream, and the blue dash line indicates that we copy the multi-task predictors pre-trained using source domain data to predict target domain tasks.}
    \label{fig:flowchart}
\end{figure*}

We now proceed to introduce our HierMUD framework that transfers the model learned from a source domain to predict multiple tasks on a target domain without any labels from the target domain in any of the tasks. In our drive-by BHM application, the two domains are vibration data and damage information collected from a vehicle passing over two structurally different bridges, and the multiple learning tasks are damage diagnostic tasks, such as damage detection, localization, and quantification. The overview flowchart of our framework is shown in Figure~\ref{fig:flowchart}. The framework contains three modules: 1) a data pre-processing module, 2) a multi-task UDA module, and 3) a target domain prediction module. In the following subsections, we present each module in detail.

\subsection{Data pre-processing module}
The data pre-processing module contains two steps: data augmentation and initial processing. In the first step, to avoid over-fitting and data biases while providing sufficient information of each class, we conduct data augmentation on raw data by multiple procedures including adding white noise, randomly cropping or erasing samples. The data augmentation expands the size of the dataset and introduces data variability, which improves the robustness of the learned multi-task UDA model.

In the second step, we create the input to our multi-task UDA module, including the source domain data with the corresponding labels and the target domain data. Feature transforms are applied to the raw input data to provide information in other feature space. For example, a Fast Fourier Transform can be used to convert the signal from its original domain (time or space) to the frequency domain. Short-Time Fourier Transform (STFT) or wavelet transform can be used to convert the time or space domain signal to the time-frequency domain. Specifically, in our drive-by BHM application, we conduct data augmentation by adding white noise to vehicle vibration signals. Then, we compute the STFT of each vertical acceleration record of the vehicle traveling over the bridge to preserve the time-frequency domain information.

\subsection{Multi-task unsupervised domain adaptation module}
\begin{figure*}[htb]
    \centering
    \includegraphics[width=1\linewidth]{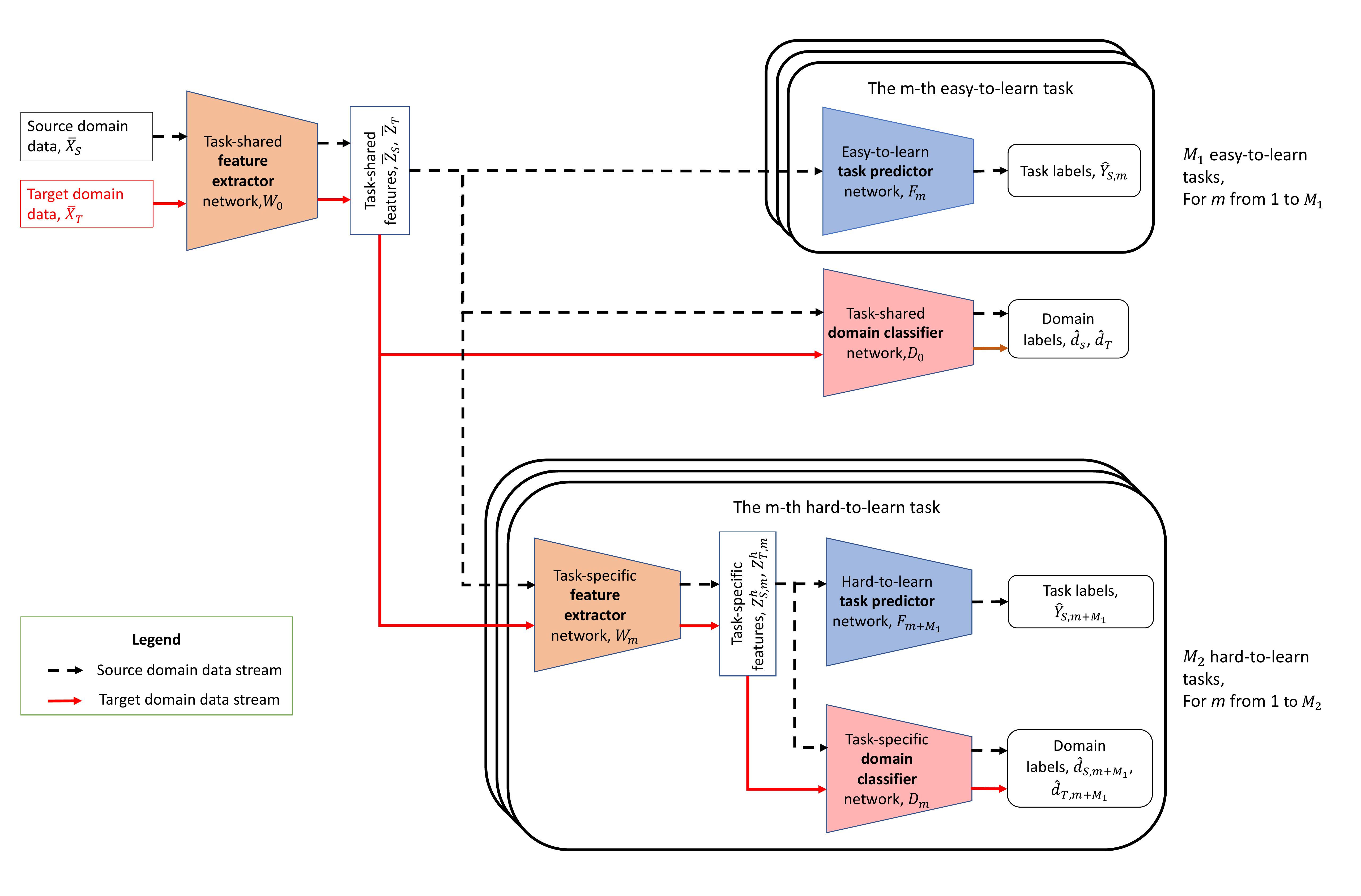}
    \caption{The architecture of our hierarchical multi-task and domain-adversarial learning algorithm. The red and black arrows between blocks represent source and target domain data stream, respectively. Orange blocks are feature extractors, blue blocks are task predictors, and red blocks are domain classifiers.}
    \label{fig:arch}
\end{figure*}

In this module, we introduce our hierarchical multi-task and domain adversarial learning algorithm (as shown in Figure~\ref{fig:arch}) that exploits the derived generalization risk bound for multi-task UDA based on the theoretical study in the previous section. This algorithm integrates domain adversarial learning and hierarchical multi-task learning to achieve an optimal trade-off between domain invariance and task informativeness. 

Our algorithm consists of three components: hierarchical feature extractors (orange blocks), task predictors (blue blocks), and domain classifiers (red blocks). Domain adversarial learning utilizes the domain classifier to minimize the domain discrepancy through an adversarial objective (e.g., Equation~\eqref{eq:divergence1}) for training against the feature extractors, which encourages the extracted feature to be domain-invariant \citep{zhang2019transfer}. The architectures of each component are presented in the following paragraphs.


\subsubsection{Architecture of hierarchical feature extractors, task predictors, and domain classifiers}~

\emph{Hierarchical feature extractors.} The hierarchical feature extractors extract domain-invariant and task-informative features. To ensure domain invariance, the parameters of the extractors are optimized with domain classifiers in an adversarial way to extract features that cannot be differentiated by the domain classifiers (i.e., domain-invariant) while the domain classifiers are optimized to best distinguish which domain the extracted features come from. 

To ensure task informativeness, we implement hierarchical feature extractors that learn task-shared and task-specific feature representations for tasks with different learning difficulties. Inspired by human learning (e.g., \citealt{kenny2012placing} and \citealt{kenny2014effectiveness}) and the work of \cite{guo2018dynamic}, we separate the total of $M$ tasks into $M_1$ easy-to-learn and $M_2$ hard-to-learn tasks based on the task difficulty, which is inversely proportional to the learning performance (e.g., prediction accuracy) in the source domain. In particular, we train, respectively, $M$ classifiers with the same model complexity using source domain data of the $M$ tasks, and obtain testing accuracy values, $\{p_1,p_2,\cdots,p_M\}$, for the $M$ tasks. A threshold $p_{t}$ is determined based on our domain knowledge and empirical observations of the problem. We then consider tasks with $p_m\geq p_t$ as easy-to-learn tasks, and tasks with $p_m\leq p_t$ as hard-to-learn tasks. For example, in our drive-by BHM application, damage detection and localization are considered as easy-to-learn tasks, and damage quantification is considered as a hard-to-learn task. 

Furthermore, we implement two types of feature extractors: one task-shared feature extractor and $M_2$ task-specific feature extractors. We denote $W_0(\cdot)$ as the task-shared feature extractor with parameters $\theta_{W_0}$, and $W_m(\cdot)$ as the task-specific feature extractor for the $m$-th hard-to-learn task with parameters $\theta_{W_m}$, where $m\in\{1,2,\cdots,M_2\}$. The source and target domain data after being processed in the data pre-processing module, ($\bar{X}_S,\bar{X}_T$), are input to the task-shared feature extractor to extract task-shared features, $\bar{Z}_S$ and $\bar{Z}_T$, for the source and target domain, respectively. For each hard-to-learn task, task-specific features, ${Z}_{S,m}^h$ and ${Z}_{T,m}^h$, are extracted from the task-shared features using the corresponding task-specific feature extractor.

The task-shared feature extractor is implemented as a deep convolutional neural network (CNN) that combines convolutional layers and pooling layers to extract feature representations. We utilize CNN to extract features because it has an excellent performance in understanding spatial hierarchies and structures of features in various resolutions \citep{goodfellow2016deep}. Further, task-specific feature extractors are implemented as deep fully-connected neural networks, consisting of multiple fully-connected layers that map task-shared features to task-specific features.

\emph{Task predictors.} Task predictors are trained to ensure that the extracted features from the hierarchical feature extractors are task-informative. They are implemented as deep fully-connected neural networks that map features to task labels. There are $M$ task predictors for all the $M$ learning tasks. We denote $F_m(\cdot)$ as the task predictor for the $m$-th task with parameters $\theta_{F_m}$, where $m\in\{1,2,\cdots,M\}$. In the training phase, the input to the task predictor of each easy-to-learn task is the task-shared feature in the source domain. The input to the task predictor of each hard-to-learn task is the corresponding task-specific feature in the source domain. Each predictor outputs the predicted source domain labels, $\hat{Y}_{S,m}$, in each task.

\emph{Domain classifiers.} Domain classifiers are trained to distinguish which domain the extracted features are from. We also have two types of domain classifiers: one task-shared domain classifier and $M_2$ task-specific domain classifiers. We denote $D_0(\cdot)$ as the task-shared domain classifier with parameters $\theta_{D_0}$, and $D_m(\cdot)$ as the task-specific domain classifier for the $m$-th hard-to-learn task with parameters $\theta_{D_m}$, where $m\in\{1,2,\cdots,M_2\}$. The task-shared domain classifier takes the task-shared features in all $M$ tasks from the source domain or the target domain as input and predicts if the feature sample comes from the source domain or not (i.e., a binary classification). Each task-specific domain classifier takes the task-specific features for each task from the source domain or the target domain as input and also classifies the feature sample into two classes (as the source or the target domain). 

We implement the domain adversarial learning by back-propagation, inspired by~\cite{ganin2016domain}, using the gradient reversal layer (GRL). We use GRL because it can be easily incorporated into any existing neural network architecture that can handle high-dimensional signals and multiple learning tasks. In particular, each domain classifier is connected to the corresponding feature extractor via a GRL that multiplies the gradient by a negative constant during back-propagation updating. With GRL, feature extractors and domain classifiers are trained in an adversarial way, such that the extracted features are as indistinguishable as possible for even well-trained domain classifiers.

Domain classifiers are implemented as deep fully-connected neural networks that map feature to domain labels. One should note that the architecture of domain classifiers is simpler than that of task predictors to avoid overusing learning resources to train domain classifiers over task predictors, which could reduce task informativeness of the extracted features \citep{ganin2016domain}.

\subsubsection{Loss function for hierarchical multi-task and domain-adversarial learning algorithm}~


In this subsection, we present the loss function for our hierarchical multi-task and domain-adversarial learning algorithm, which minimizes the objective function in Equation~\eqref{eq:loss4}. 

After considering the hierarchical structure, we can rewrite the objective function of the optimization in Equation~\eqref{eq:loss4} as

\begin{equation}
\label{eq:objective}
    \begin{aligned}
    &{\mathop{\min_{{\bf \theta}_{W_0},{\bf \theta}_{W_1}\cdots,{\bf \theta}_{W_{M_2}},}}_{{\bf \theta}_{F_{1}},\cdots,{\bf \theta}_{F_{M}},}}\Big[\sum_{m=1}^{M_1}\lambda_m\mathcal{L}_{e,m}(\theta_{W_0},\theta_{F_{m}})\\
    &+\frac{1}{M_2}\sum_{m=1}^{M_2}\lambda_{m+M_1} \mathcal{L}_{h,m}(\theta_{W_0},\theta_{W_{m}},\theta_{F_{m+M_1}})\\
    &+\lambda_{D_0}\log\sum_{m=1}^M\exp(-\min_{\theta_{D_0}}\mathcal{L}_{D_0,m}(\theta_{W_0},\theta_{D_0}))\\
    &-\min_{\theta_{D_1},\cdots,\theta_{D_M}}\sum_{m=1}^{M_2}\lambda_{D_m} \mathcal{L}_{D_m}(\theta_{W_0},\theta_{W_m},\theta_{D_m})\Big],
    \end{aligned}
\end{equation}
where $\mathcal{L}_{e,m}(\theta_{W_0},\theta_{F_{m}})
=-\mathbb{E}_{(x,y)\sim (X_{S,m},Y_{S,m})}\sum_{c=1}^{C_{m}}\mathbb{I}(y=c)\log {\bf F}_{m}({\bf W}_0(x))$
is the cross-entropy loss for the $m$-th easy-to-learn task; $C_m$ is the number of classes for the $m$-th task; $\mathcal{L}_{h,m}(\theta_{W_0},\theta_{W_{m}},\theta_{F_{m+M_1}})=-\mathbb{E}_{(x,y)\sim (X_{S,m},Y_{S,m})}\Big[\sum_{c=1}^{C_{m+M_1}}\mathbb{I}(y=c) \log {\bf F}_{m+M_1}({\bf W}_m({\bf W}_0(x)))\Big]$
is the cross-entropy loss for the $m$-th hard-to-learn tasks; $
    \mathcal{L}_{D_0,m}(\theta_{W_0},\theta_{D_0})
    =-\mathbb{E}_{x\sim X_{S,m}}[\log {\bf D}_0({\bf W}_0(x))]
    -\mathbb{E}_{x\sim X_{T,m}}[\log(1-{\bf D}_0({\bf W}_0(x)))]\big)$
is the task-shared domain classifier loss for the $m$-th tasks ($m\in{1,2,\cdots,M}$); 
$\mathcal{L}_{D_m}(\theta_{W_0},\theta_{W_m},\theta_{D_m})
    =-\mathbb{E}_{x\sim X_{S,m}}[\log {\bf D}_m({\bf W}_{m}({\bf W}_0(x)))]
    -\mathbb{E}_{x\sim X_{T,m}}[\log(1-{\bf D}_m({\bf W}_{m}({\bf W}_0(x))))]$
is the task-specific domain classifier loss; $\lambda_1,\cdots,\lambda_M$ are hyper-parameters to trade-off between easy-to-learn tasks and hard-to-learn tasks weights; and $\lambda_D,\lambda_{D_1},\cdots,\lambda_{D_{M_2}}$ are hyper-parameters to trade-off between domain-invariance and task-informative of features. 

The minimax optimization problem in Equation~\eqref{eq:objective} is solved by finding the saddle point $$\hat{\theta}_{W_0},\hat{\theta}_{W_{1}},\cdots,\hat{\theta}_{W_{M_2}},\hat{\theta}_{F_{1}},\cdots,\hat{\theta}_{F_{M}},\hat{\theta}_{D_0},\hat{\theta}_{D_1},\cdots,\hat{\theta}_{D_M},$$ such that
\begingroup
\allowdisplaybreaks
\begin{equation}
\label{eq:saddle1}
\begin{aligned}
&(\hat{\theta}_{W_0},\hat{\theta}_{W_{1}},\cdots,\hat{\theta}_{W_{M_2}},\hat{\theta}_{F_{1}},\cdots,\hat{\theta}_{F_{M}})\\&=\mathop{\argmin_{{\bf \theta}_{W_0},{\bf \theta}_{W_1}\cdots,{\bf \theta}_{W_{M_2}},}}_{{\bf \theta}_{F_{1}},\cdots,{\bf \theta}_{F_{M}}}~\Big[\frac{1}{M_1}\sum_{m=1}^{M_1}\lambda_m\mathcal{L}_{e,m}(\theta_{W_0},\theta_{F_{m}})\\
    &+\lambda_h\frac{1}{M_2}\sum_{m=1}^{M_2}\lambda_{m+M_1} \mathcal{L}_{h,m}(\theta_{W_0},\theta_{W_{m}},\theta_{F_{m+M_1}})\\
    &+\lambda_{D_0}\log\sum_{m=1}^M\exp(-\min_{\theta_{D_0}}\mathcal{L}_{D_0,m}(\theta_{W_0},\theta_{D_0}))\\
    &-\min_{\theta_{D_1},\cdots,\theta_{D_M}}\sum_{m=1}^{M_2}\lambda_{D_m} \mathcal{L}_{D_m}(\theta_{W_0},\theta_{W_m},\theta_{D_m})\Big],
\end{aligned}
\end{equation}
\begin{equation}
\label{eq:saddle2}
\begin{aligned}
\hat{\theta}_{D_0}=\argmax_{ \theta_{D_0}}\lambda_{D_0}\log\sum_{m=1}^M\exp(-\mathcal{L}_{D_0,m}(\theta_{W_0},\theta_{D_0})),
\end{aligned}
\end{equation}
\begin{equation}
\label{eq:saddle3}
\begin{aligned}
\hat{\theta}_{D_m}=\argmin_{ \theta_{D_m}}\lambda_{D_m} \mathcal{L}_{D_m}(\hat{\theta}_{W_0},\hat{\theta}_{W_m},{\theta}_{D_m}),
\end{aligned}
\end{equation}
\endgroup

A saddle point defined by Equations \eqref{eq:saddle1}-\eqref{eq:saddle3} can be found as a stationary point of the following gradient updates:
for updating hierarchical feature extractors,
\begin{equation}
\label{eq:updateW}
\begin{aligned}
\theta_{W_0}\leftarrow \theta_{W_0}-\mu\Big(&\frac{1}{M_1}\sum_{m=1}^{M_1}\lambda_m\frac{\partial\mathcal{L}_{e,m}}{\partial \theta_{W_0}}\\
&+\frac{1}{M_2}\sum_{m=1}^{M_2}\lambda_{m+M_1}\frac{\partial\mathcal{L}_{h,m}}{\partial \theta_{W_0}}\\
&-\lambda_{D_0}\sum_{m=1}^M w_m\frac{\partial\mathcal{L}_{D_0,m}}{\partial\theta_{W_0}}\\
&-\sum_{m=1}^{M_2}\lambda_{D_m}\frac{\partial\mathcal{L}_{D_m}}{\partial\theta_{W_0}}\Big),
\end{aligned}
\end{equation}
\begin{equation}
\label{eq:updateWm}
\begin{aligned}
\theta_{W_m}\leftarrow \theta_{W_m}-\mu\Big(\frac{\lambda_{m+M_1}}{M_2}\frac{\partial\mathcal{L}_{h,m}}{\partial \theta_{W_m}}-\lambda_{D_m}\frac{\partial\mathcal{L}_{D_m}}{\partial\theta_{W_m}}\Big),
\end{aligned}
\end{equation}
for updating the $m$-th easy-to-learn task predictor, where $m\in\{1,\cdots,M_1\}$,
\begin{equation}
\label{eq:updateFe}
\begin{aligned}
\theta_{F_{m}}\leftarrow \theta_{F_m}-\mu\frac{\lambda_m}{M_1}\frac{\partial\mathcal{L}_{e,m}}{\partial\theta_{F_m}},
\end{aligned}
\end{equation}
for updating the $m$-th hard-to-learn task predictor, where $m\in\{1,\cdots,M_2\}$,
\begin{equation}
\label{eq:updateFh}
\begin{aligned}
\theta_{F_{m+M_1}}\leftarrow \theta_{F_{m+M_1}}-\mu\frac{\lambda_{m+M_1}}{M_2}\frac{\partial\mathcal{L}_{e,m}}{\partial\theta_{F_{m+M_1}}},
\end{aligned}
\end{equation}
for updating domain classifiers,
\begin{equation}
\label{eq:updateD}
\begin{aligned}
\theta_{D_0}\leftarrow \theta_{D_0}-\mu\lambda_{D_0}\sum_{m=1}^Mw_m\frac{\partial\mathcal{L}_{D_0,m}}{\partial\theta_{D_0}},
\end{aligned}
\end{equation}
\begin{equation}
\label{eq:updateDm}
\begin{aligned}
\theta_{D_m}\leftarrow \theta_{D_m}-\mu\lambda_{D_m}\frac{\partial\mathcal{L}_{D_m}}{\partial\theta_{D_m}},
\end{aligned}
\end{equation}
where $\mu$ is the learning rate; $$w_m=\frac{\exp(-\min_{\theta_{D_0}}\mathcal{L}_{D_0,m}(\theta_{W_0},\theta_{D_0}))}{\sum_{m=1}^M \exp(-\min_{\theta_{D_0}}\mathcal{L}_{D_0,m}(\theta_{W_0},\theta_{D_0}))}$$ is the adaptive weight for the $m$-th task. Task with larger distribution divergence has larger weight. The updates of Equation \eqref{eq:updateFe}-\eqref{eq:updateDm} are similar to those of deep neural network models using stochastic gradient descent (SGD). For Equation \eqref{eq:updateW} and \eqref{eq:updateWm}, the difference between the updates of them and SGD updates is that the gradients from the domain classifiers are subtracted. This subtracted gradient is accomplished by inserting the aforementioned GRL between feature extractors and domain classifiers. Specifically, the forward propagation of the GRL is the same as an identity transformation, and during back-propagation, the GRL changes the sign of the gradient (i.e., multiply it by a negative constant) before passing it to the preceding layer \citep{ganin2016domain}. The optimization of our hierarchical multi-task and domain-adversarial learning algorithm is summarized in Algorithm~\ref{alg:1}.

\begin{algorithm*}
    \caption{Hierarchical multi-task and domain-adversarial learning algorithm}
\label{alg:1}
  \begin{algorithmic}[1]
    \INPUT{\begin{itemize}
        \item[-]Training iterations: $P$;
        \item[-]Batch size: $K$;
        \item[-]Number of tasks: $M$;
        \item[-]The first $M_1$ tasks are easy-to-learn tasks, the last $M_2$ tasks are hard-to-learn tasks, and $M=M_1+M_2$.
        \item[-]Number of classes for the tasks: $C_{m}$ for $m\in 1,\cdots, M$;
        \item[-]Hyper-parameters: $\lambda_1,\cdots,\lambda_M,\lambda_{D_0},\lambda_{D_1},\cdots,\lambda_{D_{M_2}}$.
        \end{itemize}}
    \OUTPUT{Neural network: $\{{\bf W}_0,{\bf D}_0,{\bf W}_{1}\cdots,{\bf W}_{M_2},{\bf F}_{1},\cdots,{\bf F}_{M},{\bf D}_{1},\cdots,{\bf D}_{M_2}\}$}
    \STATE{Randomly initialize network parameters: ${\bf \theta}_{W_0},{\bf \theta}_{D_0},{\bf \theta}_{W_1}\cdots,{\bf \theta}_{W_{M_2}},{\bf \theta}_{F_{1}},\cdots,{\bf \theta}_{F_{M}},{\bf \theta}_{D_1},\cdots,{\bf \theta}_{D_{M_2}}$}
    \FOR{$p$ from 1 to $P$}
    \STATE{\# Forward pass}
    \FOR{$m$ from 1 to $M$}
    \STATE{Sample $K$ data points and label from the source domain $\mathcal{D}_{S,m}$ for each task: $({X}_{S,m},{Y}_{S,m})$.}
    \STATE{Sample $K$ data points from the target domain $\mathcal{D}_t$ for each task: $X_{T,m}$.}
    \STATE{Compute the task-shared feature for the source domain data: $Z_{S,m}={\bf W}_0(X_{S,m}).$}
    \STATE{Compute the task-shared feature for the target domain data:  $Z_{T,m}={\bf W}_0(X_{T,m}).$}
    \ENDFOR
    \FOR{$m$ from 1 to $M_2$}
    \STATE{Compute the task-specific feature for the source domain hard-to-learn tasks: $Z_{S,m}^h={\bf W}_m(Z_{S,m+M_1}).$}
    \STATE{Compute the task-specific feature for the target domain hard-to-learn tasks: $Z_{S,m}^h={\bf W}_m(Z_{T,m+M_1}).$}
    \ENDFOR
    
    \STATE{\# Backward pass}
    \STATE{Update the feature extractors and task predictors using equation~\eqref{eq:objective}: \begin{align*}(\hat{\theta}_{W_0},\hat{\theta}_{W_{1}},\cdots,\hat{\theta}_{W_{M_2}},\hat{\theta}_{F_{1}},\cdots,\hat{\theta}_{F_{M}})&\\=\mathop{\argmin_{{\bf \theta}_{W_0},{\bf \theta}_{W_1}\cdots,{\bf \theta}_{W_{M_2}},}}_{{\bf \theta}_{F_{1}},\cdots,{\bf \theta}_{F_{M}}}~\Big[&\frac{1}{M_1}\sum_{m=1}^{M_1}\lambda_m\mathcal{L}_{e,m}(\theta_{W_0},\theta_{F_{m}})+\frac{1}{M_2}\sum_{m=1}^{M_2}\lambda_{m+M_1} \mathcal{L}_{h,m}(\theta_{W_0},\theta_{W_{m}},\theta_{F_{m+M_1}})\\
    &+\lambda_{D_0}\log\sum_{m=1}^M\exp(-\min_{\theta_{D_0}}\mathcal{L}_{D_0,m}(\theta_{W_0},\theta_{D_0}))\\
    &-\min_{\theta_{D_1},\cdots,\theta_{D_M}}\sum_{m=1}^{M_2}\lambda_{D_m} \mathcal{L}_{D_m}(\theta_{W_0},\theta_{W_m},\theta_{D_m})\Big],
    \end{align*}}
    \STATE{Update the task-shared domain classifier: $\hat{\theta}_{D_0}=\argmax_{ \theta_{D_0}}\lambda_{D_0}\log\sum_{m=1}^M\exp(-\mathcal{L}_{D_0,m}(\theta_{W_0},\theta_{D_0})).$}
    \FOR{$m$ in $1:M_2$}
    \STATE{Update the task-specific domain classifier: $\hat{\theta}_{D_m}=\argmin_{ \theta_{D_m}}\lambda_{D_m} \mathcal{L}_{D_m}(\hat{\theta}_{W_0},\hat{\theta}_{W_m},{\theta}_{D_m}).$}
    \ENDFOR
    \ENDFOR
  \end{algorithmic}
\end{algorithm*}

\subsection{Target domain prediction module}
\begin{figure}[htb]
    \centering
    \includegraphics[width=0.95\linewidth]{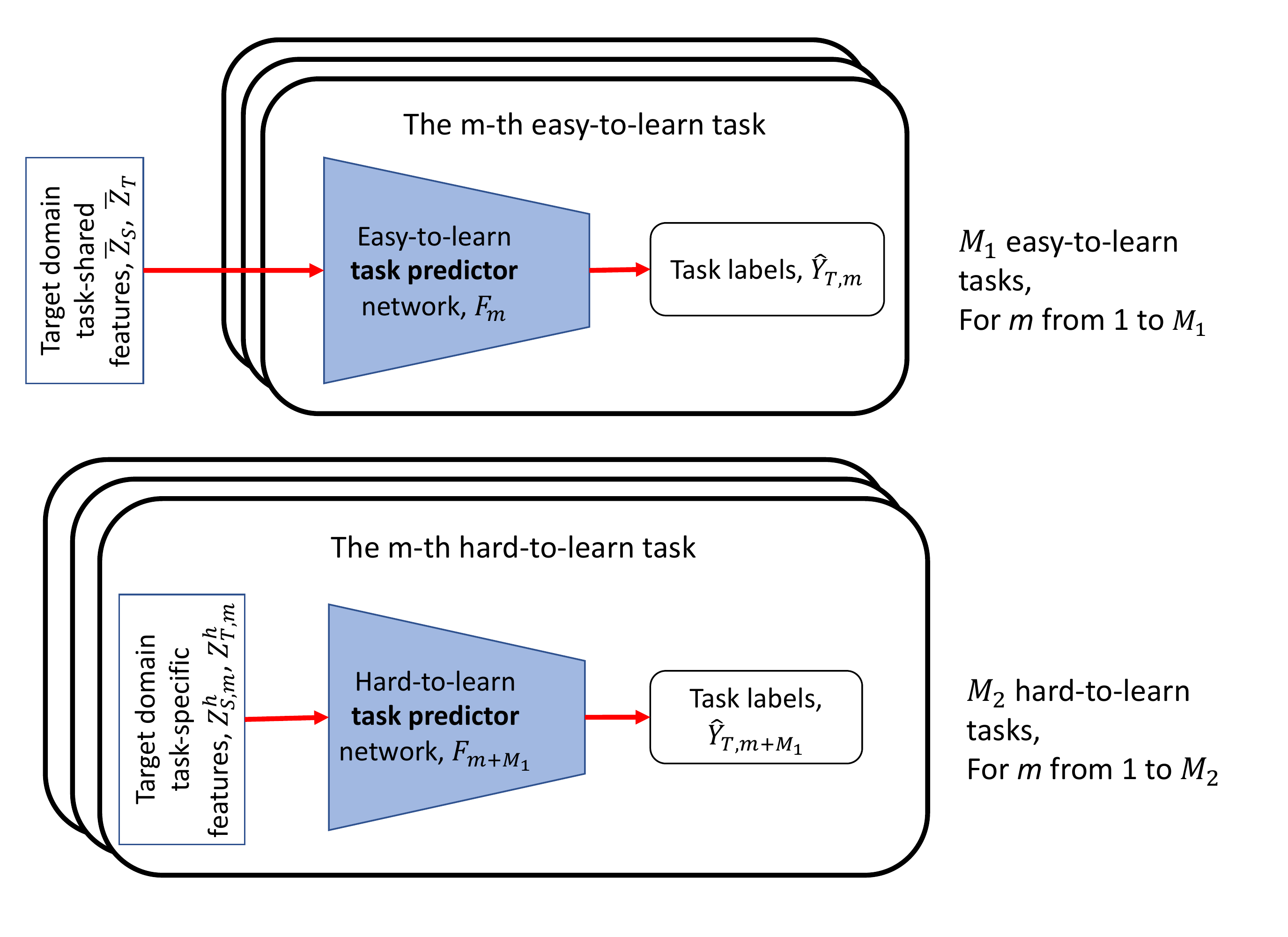}
    \caption{The architecture of our hierarchical multi-task and domain-adversarial learning algorithm in the target domain prediction module.}
    \label{fig:arch_prediction}
\end{figure}

 The architecture of the target domain prediction module is shown in Figure~\ref{fig:arch_prediction}. In this module, the extracted target domain task-shared features are input to the pre-trained easy-to-learn predictors to predict target domain labels in the easy-to-learn tasks, and the extracted target domain task-specific features are input to the pre-trained hard-to-learn predictors to predict target domain labels in the hard-to-learn tasks.

\section{Evaluation}
In this section, we evaluate our HierMUD framework for drive-by BHM using data collected from lab-scale experiments. The experiments are conducted on two structurally different bridges using three vehicles of different weights.

\subsection{Experimental setup and data description}
\begin{figure*}[htb]
    \centering
    \begin{minipage}{0.35\textwidth}
        \centering
        \includegraphics[width=1\linewidth]{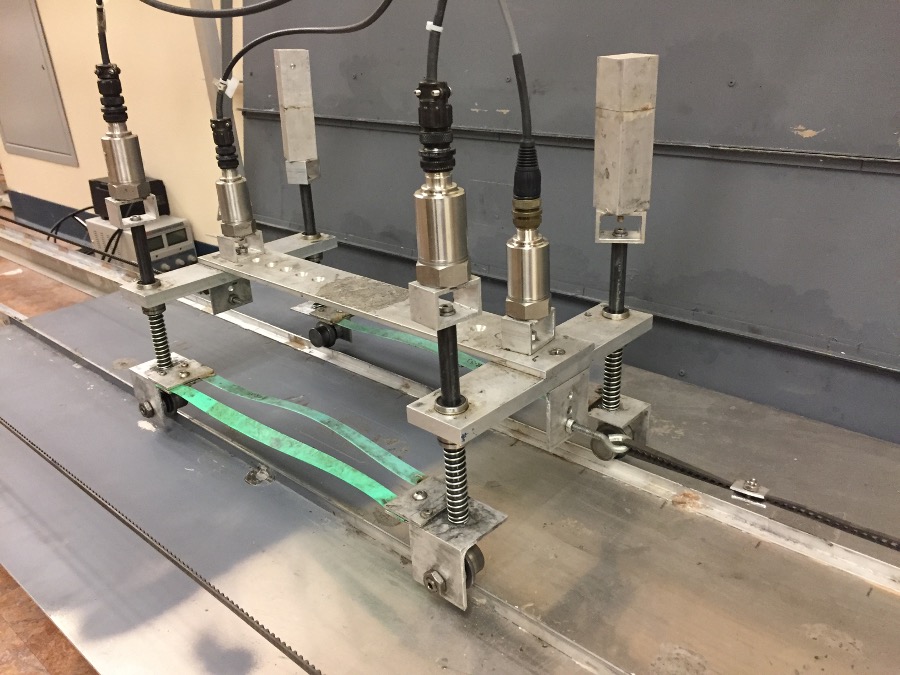}
        {(a)}
        \label{fig:veh}
    \end{minipage}\qquad
    \begin{minipage}{0.35\textwidth}
        \centering
        \includegraphics[width=1\linewidth]{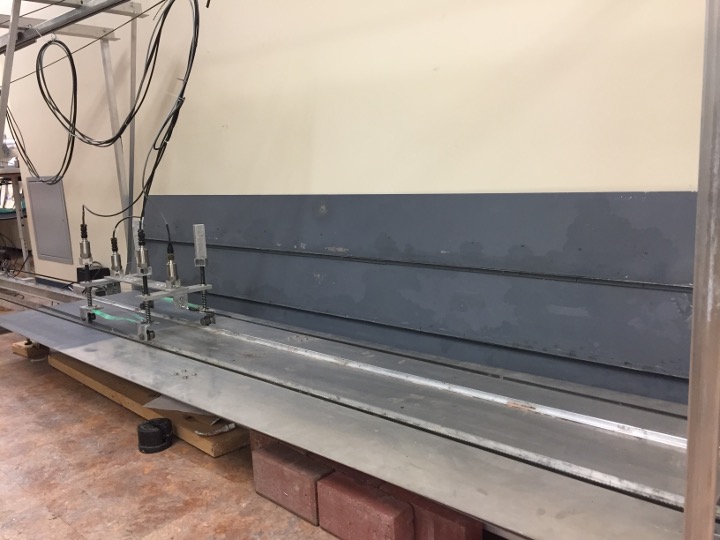}
        {(b)}
        \label{fig:bri}
    \end{minipage}
    \caption{(a) A vehicle (V1) moving at a controlled speed and (b) a bridge (B1) that the vehicle passes.}
    \label{fig:exp_setup}
\end{figure*}

A lab-scale VBI system, as shown in Figure~\ref{fig:exp_setup}, was employed to create the dataset. The experiments involved two 8-foot bridges (B1 and B2) with different weights of 34.2 lb and 43.0 lb, different dominant frequencies of 5.9 Hz and 7.7 Hz, and damping ratios of 0.13 and 0.07, respectively. The data are collected from three small-scale vehicles (V1, V2, and V3) with different weights of 10.6 lb, 11.6 lb, and 12.6 lb, respectively, that were driven over the bridge. Vertical acceleration signals were collected from four accelerometers mounted on each vehicle (front chassis, back chassis, front wheel, and back wheel) while they moved individually across the bridge at a constant speed (0.75 m/s). The sampling rate of all sensors is 1600 Hz.

Damage proxy is introduced by adding mass at different locations of the bridge. For varying damage severity, the magnitude of the attached mass for each run ranged from 0.5 lb to 2.0 lb with an interval of 0.5 lb. A heavier mass means more severe damage since it induces more significant structural change from the initial condition (i.e., healthy state). Each damage severity level was induced at three different damage locations ($l$ is every quarter of the bridge span). For each damage severity and location scenario for each vehicle and bridge combination, the experiments were repeated thirty times (i.e., 30 trials of a vehicle passing a bridge). In total, the dataset includes 2 (bridges) $\times$ 3 (vehicles) $\times$ [3 (damage locations) $\times$ 4 (damage severity levels) + 1 (undamaged case)] $\times$ 30 (iterations) = 2340 (trials), which results in 2340 (trials) $\times$ 4 (sensors) = 9360 (records). Details of the experimental instrumentation can be found in \citep{liu2020diagnosis}. 

In summary, our drive-by BHM problem has three tasks: binary damage detection, 3-class damage localization, and 4-class damage quantification. For each damage diagnostic task, two model transfers, from B1 to B2 and from B2 to B1, using signals collected from each of the three vehicles (V1, V2, and V3), are conducted, making a total of six evaluations.

\subsection{Setup of our HierMUD framework}

\begin{figure*}[htb]
    \centering
    \includegraphics[width=0.9\linewidth]{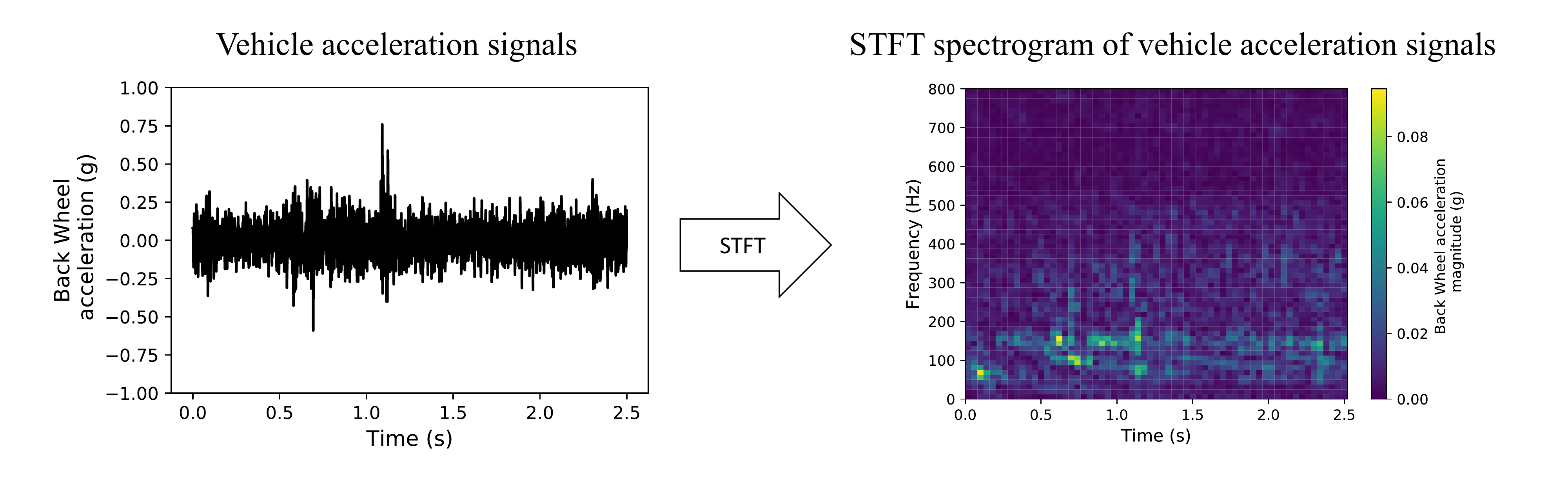}
    \caption{An example of the raw acceleration signal collected from a vehicle passing over a lab-scale bridge and its corresponding STFT representation.}
    \label{fig:signal}
\end{figure*}

In this subsection, we describe our data preprocessing procedure and the setup of our HierMUD framework. We first preprocessed the input data by conducting data augmentation. Data augmentation adds white noise to vehicle acceleration signals. The white noise has zero mean and variance of mean squared magnitude of the vehicle acceleration signal. Then, we compute the short-time Fourier transform (STFT) representation of each signal to preserve the time-frequency domain information. 
The size of each input data is $C\times W\times H$, where $C$ is the number of sensor channels on the vehicle, which is 4 in our system; $W$ and $H$ are respectively the number of time segments and the sample frequencies of the STFT representation. Figure~\ref{fig:signal} shows an example of the vehicle acceleration signal and its corresponding STFT representation.

\begin{figure*}[htb]
    \centering
    \includegraphics[width=1\linewidth]{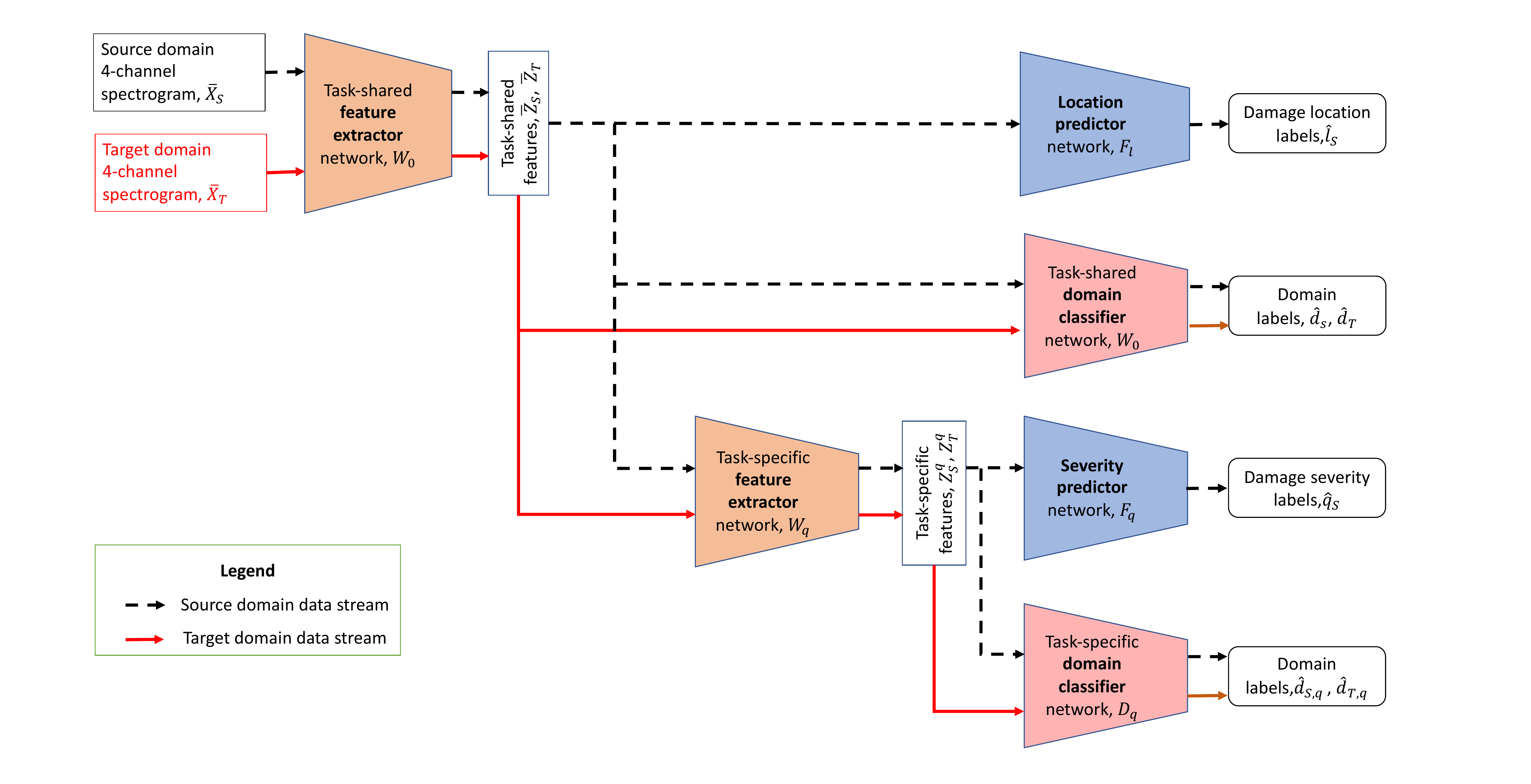}
    \caption{The architecture of our multi-task and domain adversarial learning algorithm applied on the drive-by BHM.}
    \label{fig:arch_bridge}
\end{figure*}

We consider damage detection and localization as easy-to-learn tasks and damage quantification as a hard-to-learn task. This is because we obtained higher supervised prediction accuracy for damage detection and localization tasks than that for the quantification task, and we observed that the data distributions for different damage locations are more separable than that for different damage severity levels. We then develop the hierarchical architecture that extracts task-shared features for all the tasks and further task-specific features  for the quantification task. Moreover, the existence of damage (for damage detection task) is represented by introducing an additional label within the location predictor, instead of creating an additional binary damage detection classifier that increases the model complexity. After all, the overall architecture of the neural network modules used in our HierMUD model are shown in Figure~\ref{fig:arch_bridge} and Table~\ref{tab:detailed_arch}. 


\begin{table*}[htb]
  \centering
  \begin{tabular}{llccc}
  \hline
  Network & Layer & Patch Size & Input Size & Activation\\
    \hline
    Task-shared feature extractor & Convolution (2D) & $64\times 5\times 5$ & $4\times 64\times 64$ & \\
    & Max Pooling & $2\times 2$ & $64\times 60\times 60$ &LeakyReLU\\
    & Convolution (2D) & $50\times 5\times 5$ & $64\times 30\times 30$ &  \\
    & Max Pooling & $2\times 2$ & $64\times 26\times 26$ &LeakyReLU\\
    & Convolution (2D) & $50\times 3\times 3$ & $50\times 13\times 13$ & \\
    & Max Pooling & $2\times 2$ & $50\times 11\times 11$ & LeakyReLU\\
    \hline
    Task-specific feature extractor& Flatten & & $50\times 5\times 5$&\\
    & Full connection &  $1250\times 1250$ & $1250\times 1$& ReLU\\
    \hline
    Location predictors& Flatten & & $50\times 5\times 5$&\\
    & Full connection & $1250\times 100$ & $100\times 1$& ReLU\\
    & Full connection & $100\times 4$ & $100\times 1$& Softmax\\
    \hline
    Severity predictors& Flatten & & $50\times 5\times 5$&\\
    & Full connection & $1250\times 100$ & $100\times 1$& ReLU\\
    & Full connection & $100\times 5$ & $100\times 1$& Softmax\\
    \hline
    Task-shared and task-specific domain classifiers& Flatten & & $50\times 5\times 5$&\\
    & Full connection & $1250\times 2$ & $1250\times 1$& Softmax\\
    \hline
  \end{tabular}
  \caption{Detailed network architectures of HierMUD.}
  \label{tab:detailed_arch}
\end{table*}

\subsection{Baseline methods}
We compare the performance of HierMUD with five baseline methods described below:
\begin{itemize}
    \item[1)] MCNN is a multi-task convolutional neural network model that directly applies the model trained using source domain data to the target bridge without applying any domain adaptation. 
    \item[2)] iUD is an independent task learning model with UDA. It predicts damage location and severity levels using two independent UDA models.
    \item[3)] sUD is a sequential task learning model with UDA. It predicts damage location and severity level step-by-step instead of using multi-task learning.
    \item[4)] MUD directly combines MTL and UDA. It predicts damage location and severity level simultaneously, but does not use the hierarchical structure and the soft-max objective.
    \item[5)] HierMUD-a further adds the hierarchical structure to MUD, but it optimizes the average objective - Equation~\ref{eq:loss3}.
\end{itemize}
The differences of these methods are also summarized in Table~\ref{tab:baseline}. Note that to have a fair comparison, components (feature extractors, task predictors, and domain classifiers) of each method have the same architecture as HierMUD method.
\begin{table*}[htb]
  \centering
  \begin{tabular}{l|c|c|c|c|c|c}
  \hline
    Method & MTL & ITL & STL & UDA & Hierarchical structure & Soft-max objective \\
    \hline
    MCNN & \checkmark & & & & & \\
    iUD & & \checkmark & & \checkmark & & \\
    sUD & & & \checkmark & \checkmark & & \\
    MUD & \checkmark & & & \checkmark & & \\
    HierMUD-a & \checkmark & & & \checkmark & \checkmark & \\
    HierMUD (ours) & \checkmark & & & \checkmark & \checkmark & \checkmark \\
    \hline
  \end{tabular}
  \caption{The comparison of our method and baseline methods. MTL, ITL, STL, and UDA stand for multi-task learning, independent task learning, sequential task learning, and unsupervised domain adaptation, respectively.}
  \label{tab:baseline}
\end{table*}

\subsection{Unsupervised hyper-parameter selection}

In this section, we describe an approach to fine-tune hyper-parameters (including the learning rate, network architecture, and trade-off hyper-parameters) for performing our multi-task and domain-adversarial learning algorithm. Because there is no labeled data in the target domain during the training phase, we select hyper-parameters by conducting a reverse validation~\citep{zhong2010cross,ganin2016domain}. In short, the labeled source samples $\bar{X}_{S}$ and unlabeled target samples $\bar{X}_T$ are split into training sets ($\bar{X}'_S$ and $\bar{X}'_T$) which contain 90\% of $\bar{X}_S$ and $\bar{X}_T$ and the validation sets ($\bar{X}_S^V$ and $\bar{X}_S^V$), where $\bar{X}' \cup \bar{X}^V = \bar{X}$ and they are mutually exclusive. The training sets $\bar{X}'_S$ and $\bar{X}'_T$ are used to learn a target domain prediction model $\eta(\cdot)$. We then learn another prediction model $\eta_r(\cdot)$ using unlabeled target domain data, $\bar{X}'_T$, with the predicted labels ($\eta(x)$ for $x\in\bar{X}'_T$). The reverse classifier $\eta_r(\cdot)$ is evaluated on the source domain validation set $\bar{X}_S^V$. This evaluation process is conducted for several times with different values of hyper-parameters. We conduct 10-fold cross-validation with each set of hyper-parameters and choose the hyper-parameters that give the highest cross-validation accuracy for the reverse classifier on the source domain validation set.

\section{Results and discussion}
In this section, we present our evaluation results and discuss our findings. Table~\ref{tab:results} presents the results for knowledge transfer from B1 to B2 and from B2 to B1 using each of the three vehicles' data. For the binary damage detection, we report F1-scores as the performance metric because the numbers of damaged and undamaged data samples are imbalanced, and for the 3-class damage localization and 4-class quantification, we report classification accuracy (the number of correct predictions divided by the total number of predictions made) as the performance metric. There are six evaluations in total (for 2 bridges and 3 vehicles), and each evaluation was conducted 10 times with different random seeds for splitting train and validation sets and initializing model parameters. Therefore, we have total 60 tests. Each column in Table~\ref{tab:results} shows the performance for each of the six evaluations. The last column of the table provides the overall performance of all six evaluations. The numbers in each cell of the table are: average performance ($\pm$ 95\% confidence interval). 

\subsection{Performance comparison with baseline methods}

Except for the damage detection task for the evaluation using V1 response to transfer model from B2 to B1, our method (HierMUD) outperforms other baselines. Overall, HierMUD has the best damage localization and quantification accuracy, which are 93\% and 48\% on average, respectively. The average objective (HierMUD-a) has the best damage detection F1-score (96\%), which is one percent better than HierMUD. HierMUD-a and HierMUD are about twice as good as the baseline without domain adaptation (MCNN) in all three diagnostic tasks and 1.5 times as good as other baselines (iUD, sUD, and MUD) in the hard-to-learn quantification task. HierMUD achieves the best 99\% average F1-score (up to 100\% in the best test) in the damage detection task using signals collected from V2 for model transfer from B1 to B2, the best 98\% average accuracy (up to 100\% in the best test) in the damage localization task, and the best 59\% average accuracy (up to 72\% in the best test) in the damage quantification task using signals collected from V3 for model transfer from B1 to B2.

\begin{table*}[htb]
  \centering
  \resizebox{\linewidth}{!}{%
  \begin{tabular}{llccccccc}
  \hline
    Task & \multirow{2}{*}{Method} &\multicolumn{2}{c}{Vehicle 1} &\multicolumn{2}{c}{Vehicle 2} &\multicolumn{2}{c}{Vehicle 3} &\multirow{2}{*}{Overall} \\\cline{3-8}
    (Metric) & & B1$\to$B2 & B2$\to$B1 & B1$\to$B2 & B2$\to$B1 & B1$\to$B2 & B2$\to$B1 & \\
    \hhline{=========} 
     & MCNN &{\footnotesize $0.46 (\pm 0.02)$} &{\footnotesize $0.54(\pm 0.08)$} &{\footnotesize $0.45(\pm 0.05)$} &{\footnotesize $0.53(\pm 0.07)$} &{\footnotesize $0.44(\pm 0.04)$} &{\footnotesize $0.51(\pm 0.08)$} &{\footnotesize $0.49(\pm 0.03)$}\\
    \cline{2-9}
    Damage & iUD &{\footnotesize $0.57(\pm 0.16)$} &{\footnotesize $0.83(\pm 0.19)$} &{\footnotesize $0.61(\pm 0.19)$} &{\footnotesize $0.98(\pm 0.01)$} &{\footnotesize $0.94(\pm 0.07)$} &{\footnotesize $0.84(\pm 0.19)$} &{\footnotesize $0.80(\pm 0.08)$}\\
    \cline{2-9}
    detection & sUD &{\footnotesize $0.53(\pm 0.15)$} &{\footnotesize $\mathbf{0.97(\pm 0.03)})$} &{\footnotesize $0.76(\pm 0.20)$} &{\footnotesize $0.97(\pm 0.03)$} &{\footnotesize $0.76(\pm 0.20)$} &{\footnotesize $0.84(\pm 0.19)$} &{\footnotesize $0.81(\pm 0.07)$}\\
    \cline{2-9}
    (F1-score) & MUD &{\footnotesize $0.91(\pm 0.05)$} &{\footnotesize $0.92(\pm 0.02)$} &{\footnotesize $0.98(\pm 0.02)$} &{\footnotesize $0.95(\pm 0.03)$} &{\footnotesize $0.92(\pm 0.07)$} &{\footnotesize $0.95(\pm 0.02)$} &{\footnotesize $0.94(\pm 0.02)$}\\
    \cline{2-9}
     & HierMUD-a &{\footnotesize $\mathbf{0.94(\pm 0.06})$} &{\footnotesize $0.93(\pm 0.04)$} &{\footnotesize $\mathbf{0.99(\pm 0.01)}$} &{\footnotesize $0.94(\pm 0.03)$} &{\footnotesize $0.95(\pm 0.03)$} &{\footnotesize $\mathbf{0.97(\pm 0.02})$} &{\footnotesize $\mathbf{0.96(\pm 0.01)}$}\\
    \cline{2-9}
     & HierMUD &{\footnotesize $0.92(\pm 0.04)$} &{\footnotesize $0.92(\pm 0.05)$} &{\footnotesize $\mathbf{0.99(\pm 0.01)}$} &{\footnotesize $\mathbf{0.98(\pm 0.01})$} &{\footnotesize $\mathbf{0.96(\pm 0.02})$} &{\footnotesize ${0.96(\pm 0.03})$} &{\footnotesize $0.95(\pm 0.01)$}\\
    \hhline{=========} 
     & MCNN &{\footnotesize $0.35(\pm 0.05)$} &{\footnotesize $0.34(\pm 0.04)$} &{\footnotesize $0.38(\pm 0.07)$} &{\footnotesize $0.36(\pm 0.07)$} &{\footnotesize $0.40(\pm 0.04)$} &{\footnotesize $0.26(\pm 0.02)$} &{\footnotesize $0.35(\pm 0.02)$}\\
    \cline{2-9}
    Damage & iUD &{\footnotesize $0.62(\pm 0.13)$} &{\footnotesize $0.58(\pm 0.21)$} &{\footnotesize $0.61(\pm 0.15)$} &{\footnotesize $0.92(\pm 0.15)$} &{\footnotesize $0.90(\pm 0.09)$} &{\footnotesize $0.53(\pm 0.20)$} &{\footnotesize $0.69(\pm 0.08)$}\\
    \cline{2-9}
    localization & sUD &{\footnotesize $0.70(\pm 0.08)$} &{\footnotesize $0.72(\pm 0.24)$} &{\footnotesize $0.49(\pm 0.14)$} &{\footnotesize $0.72(\pm 0.24)$} &{\footnotesize $0.86(\pm 0.11)$} &{\footnotesize $0.59(\pm 0.21)$} &{\footnotesize $0.68(\pm 0.08)$}\\
    \cline{2-9}
    (Accuracy) & MUD &{\footnotesize $0.85(\pm 0.12)$} &{\footnotesize $0.86(\pm 0.04)$} &{\footnotesize $0.84(\pm 0.10)$} &{\footnotesize $0.87(\pm 0.07)$} &{\footnotesize $0.95(\pm 0.04)$} &{\footnotesize $0.95(\pm 0.02)$} &{\footnotesize $0.89(\pm 0.03)$}\\
    \cline{2-9}
     & HierMUD-a &{\footnotesize $0.89(\pm 0.05)$} &{\footnotesize $\mathbf{0.88(\pm 0.05})$} &{\footnotesize $0.82(\pm 0.11)$} &{\footnotesize $0.91(\pm 0.04)$} &{\footnotesize $0.97(\pm 0.03)$} &{\footnotesize $0.90(\pm 0.09)$} &{\footnotesize $0.90(\pm 0.03)$}\\
    \cline{2-9}
     & HierMUD &{\footnotesize $\mathbf{0.92(\pm 0.03})$} &{\footnotesize $0.87(\pm 0.05)$} &{\footnotesize $\mathbf{0.92(\pm 0.06})$} &{\footnotesize $\mathbf{0.93(\pm 0.04})$} &{\footnotesize $\mathbf{0.98(\pm 0.01})$} &{\footnotesize $\mathbf{0.95(\pm 0.04})$} &{\footnotesize $\mathbf{0.93(\pm 0.02)}$}\\
    \hhline{=========} 
     & MCNN &{\footnotesize $0.31(\pm 0.02)$} &{\footnotesize $0.31(\pm 0.01)$} &{\footnotesize $0.31(\pm 0.01)$} &{\footnotesize $0.30(\pm 0.02)$} &{\footnotesize $0.31(\pm 0.01)$} &{\footnotesize $0.28(\pm 0.02)$} &{\footnotesize $0.30(\pm 0.01)$}\\
    \cline{2-9}
    Damage & iUD &{\footnotesize $0.31(\pm 0.02)$} &{\footnotesize $0.31(\pm 0.03)$} &{\footnotesize $0.33(\pm 0.02)$} &{\footnotesize $0.35(\pm 0.03)$} &{\footnotesize $0.32(\pm 0.02)$} &{\footnotesize $0.29(\pm 0.02)$} &{\footnotesize $0.32(\pm 0.01)$}\\
    \cline{2-9}
    quantification & sUD &{\footnotesize $0.32(\pm 0.02)$} &{\footnotesize $0.34(\pm 0.01)$} &{\footnotesize $0.34(\pm 0.02)$} &{\footnotesize $0.34(\pm 0.04)$} &{\footnotesize $0.33(\pm 0.02)$} &{\footnotesize $0.29(\pm 0.02)$} &{\footnotesize $0.33(\pm 0.01)$}\\
    \cline{2-9}
    (Accuracy) & MUD&{\footnotesize $0.31(\pm 0.05)$} &{\footnotesize $0.30(\pm 0.06)$} &{\footnotesize $0.40(\pm 0.07)$} &{\footnotesize $0.33(\pm 0.07)$} &{\footnotesize $0.40(\pm 0.07)$} &{\footnotesize $0.37(\pm 0.07)$} &{\footnotesize $0.35(\pm 0.03)$}\\
    \cline{2-9}
     & HierMUD-a&{\footnotesize $0.35(\pm 0.04)$} &{\footnotesize $\mathbf{0.39(\pm 0.05})$} &{\footnotesize $0.45(\pm 0.05)$} &{\footnotesize $0.43(\pm 0.05)$} &{\footnotesize $0.46(\pm 0.08)$} &{\footnotesize $0.49(\pm 0.06)$} &{\footnotesize $0.43(\pm 0.03)$}\\
    \cline{2-9}
     & HierMUD&{\footnotesize $\mathbf{0.45(\pm 0.07})$} &{\footnotesize $\mathbf{0.39(\pm 0.05})$} &{\footnotesize $\mathbf{0.51(\pm 0.04})$} &{\footnotesize $\mathbf{0.44(\pm 0.05})$} &{\footnotesize $\mathbf{0.59(\pm 0.08})$} &{\footnotesize $\mathbf{0.51(\pm 0.06})$} &{\footnotesize $\mathbf{0.48(\pm 0.03)}$}\\
    \hline
  \end{tabular}
  }
  \caption{The performance of baseline methods and our method on the lab-scale VBI dataset. "A $\to$ B" indicates that we transfer knowledge from the source domain A to the target domain B. The numbers in each cell are: average performance (confidence interval). We bold the number that has the best result in each task.}
  \label{tab:results}
\end{table*}

\begin{figure*}[htb]
    \centering
    \begin{minipage}{0.33\textwidth}
        \centering
        \includegraphics[width=1\linewidth]{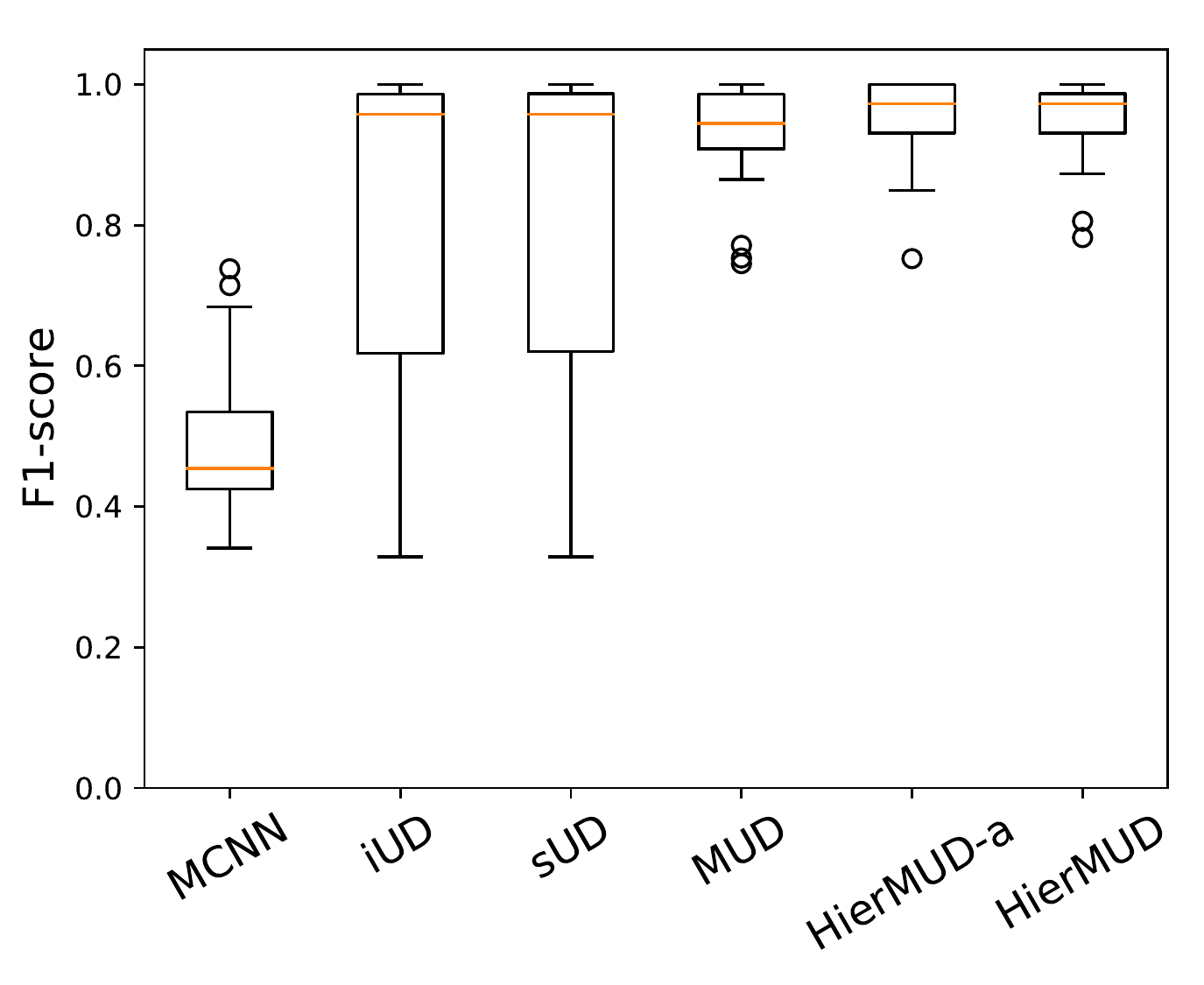}
        {(a) Overall}
        \label{fig:d}
    \end{minipage}%
    \begin{minipage}{0.33\textwidth}
        \centering
        \includegraphics[width=1\linewidth]{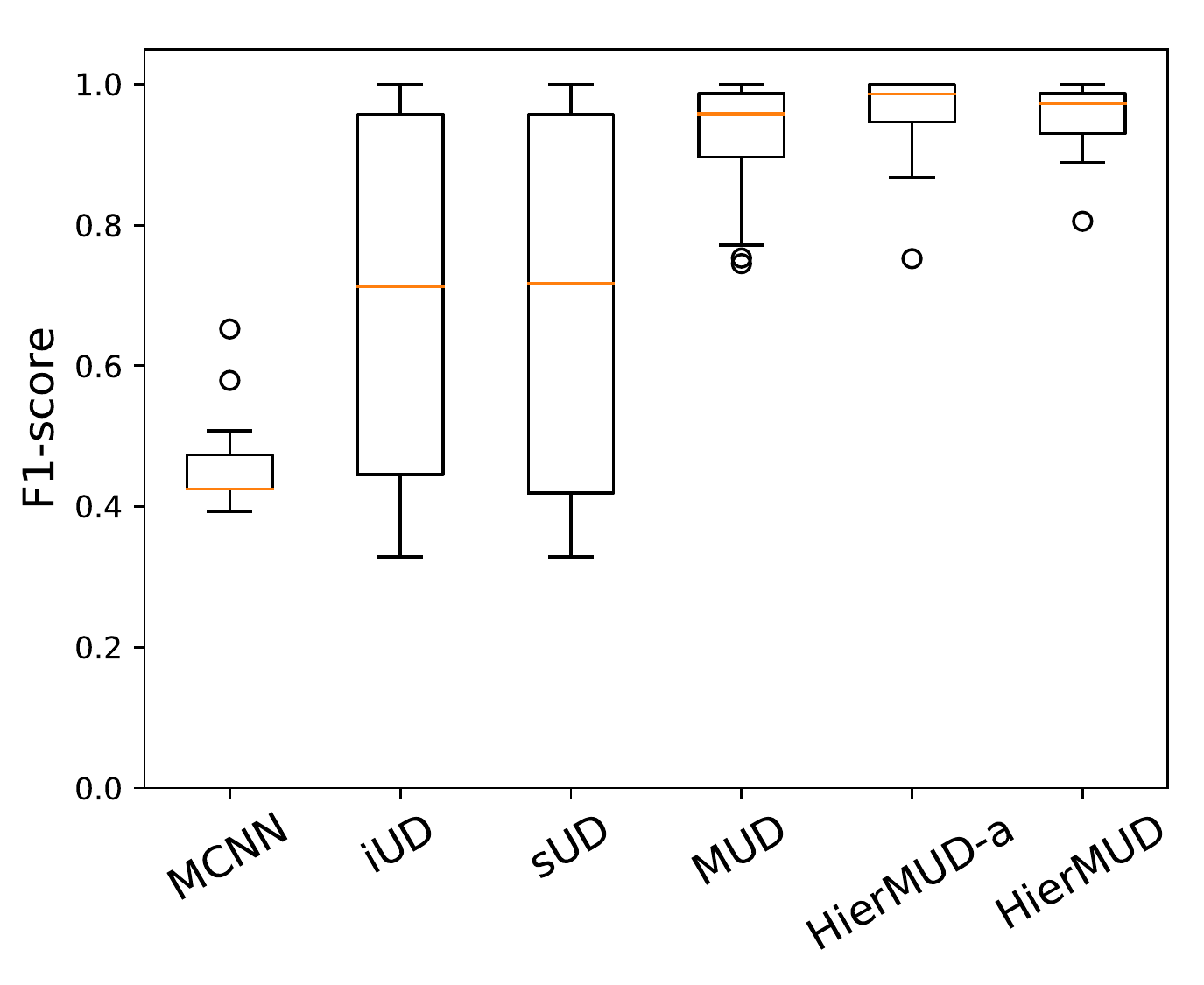}
        {(b) B1$\to$B2}
        \label{fig:df}
    \end{minipage}
    \begin{minipage}{0.33\textwidth}
        \centering
        \includegraphics[width=1\linewidth]{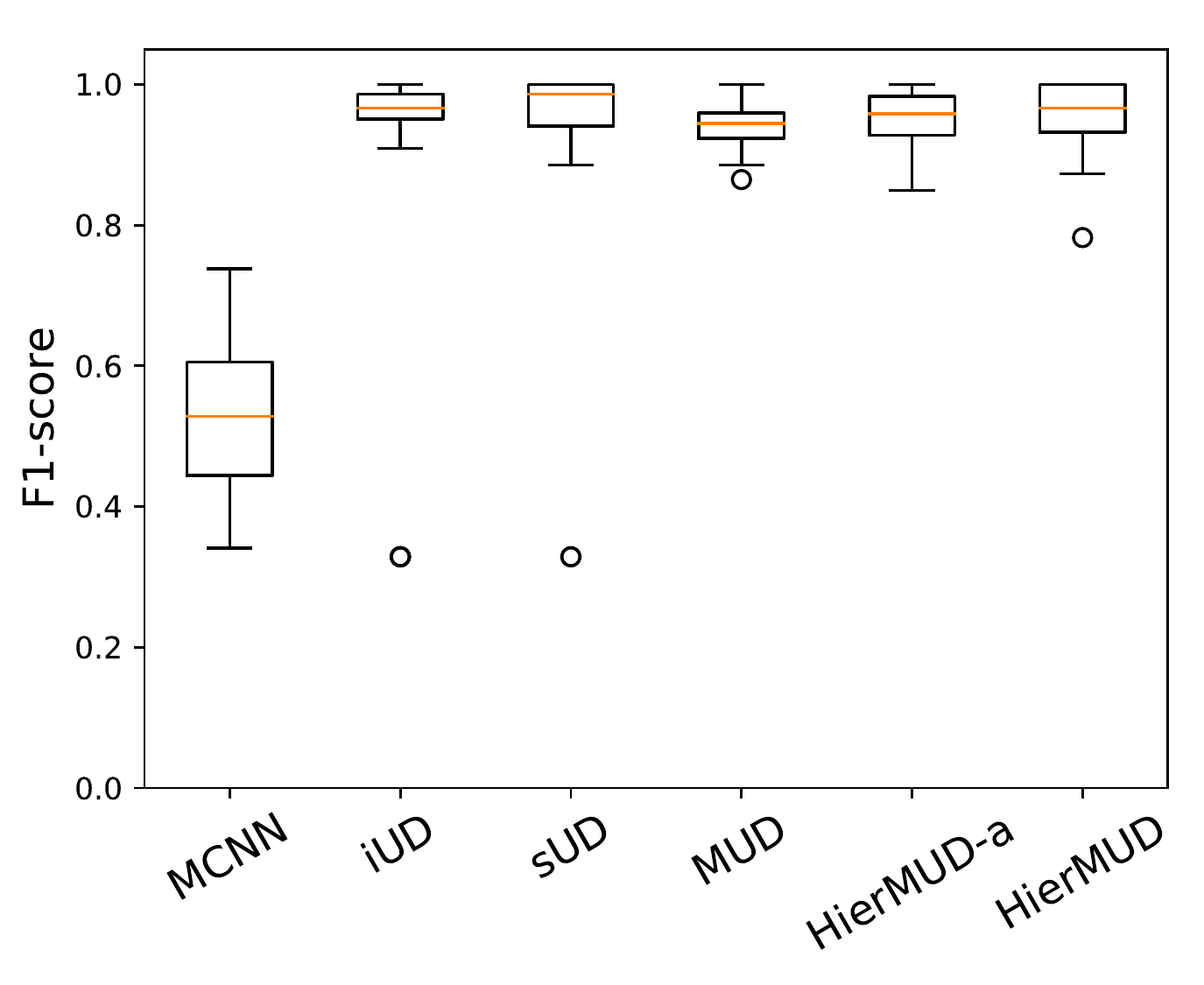}
        {(c) B2$\to$B1}
        \label{fig:db}
    \end{minipage}
    \caption{The performance of baseline methods and our method for damage detection task on the lab-scale VBI dataset.}
    \label{fig:det}
\end{figure*}

\begin{figure*}[htb]
    \centering
    \begin{minipage}{0.33\textwidth}
        \centering
        \includegraphics[width=1\linewidth]{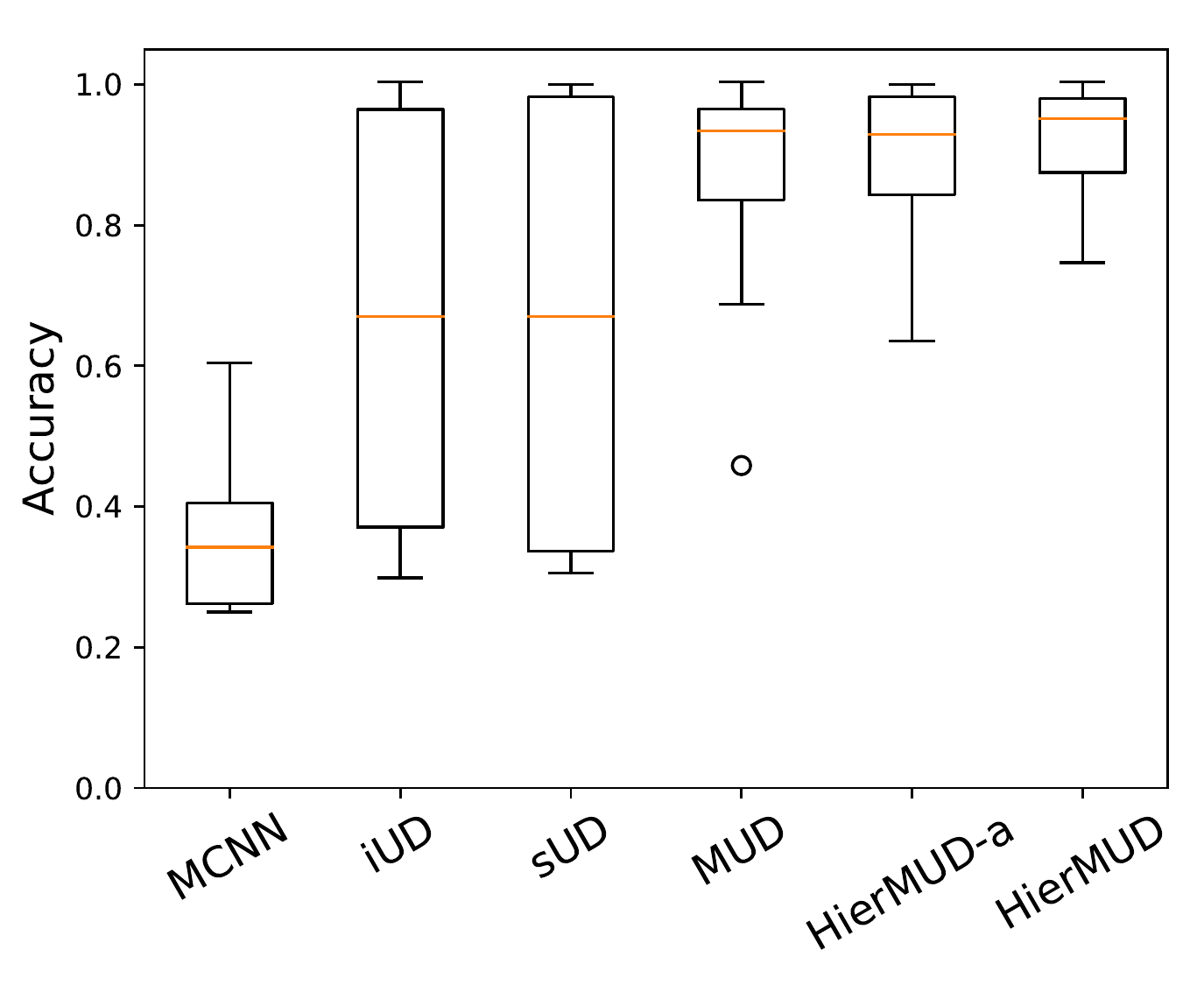}
        {(a) Overall}
        \label{fig:l}
    \end{minipage}%
    \begin{minipage}{0.33\textwidth}
        \centering
        \includegraphics[width=1\linewidth]{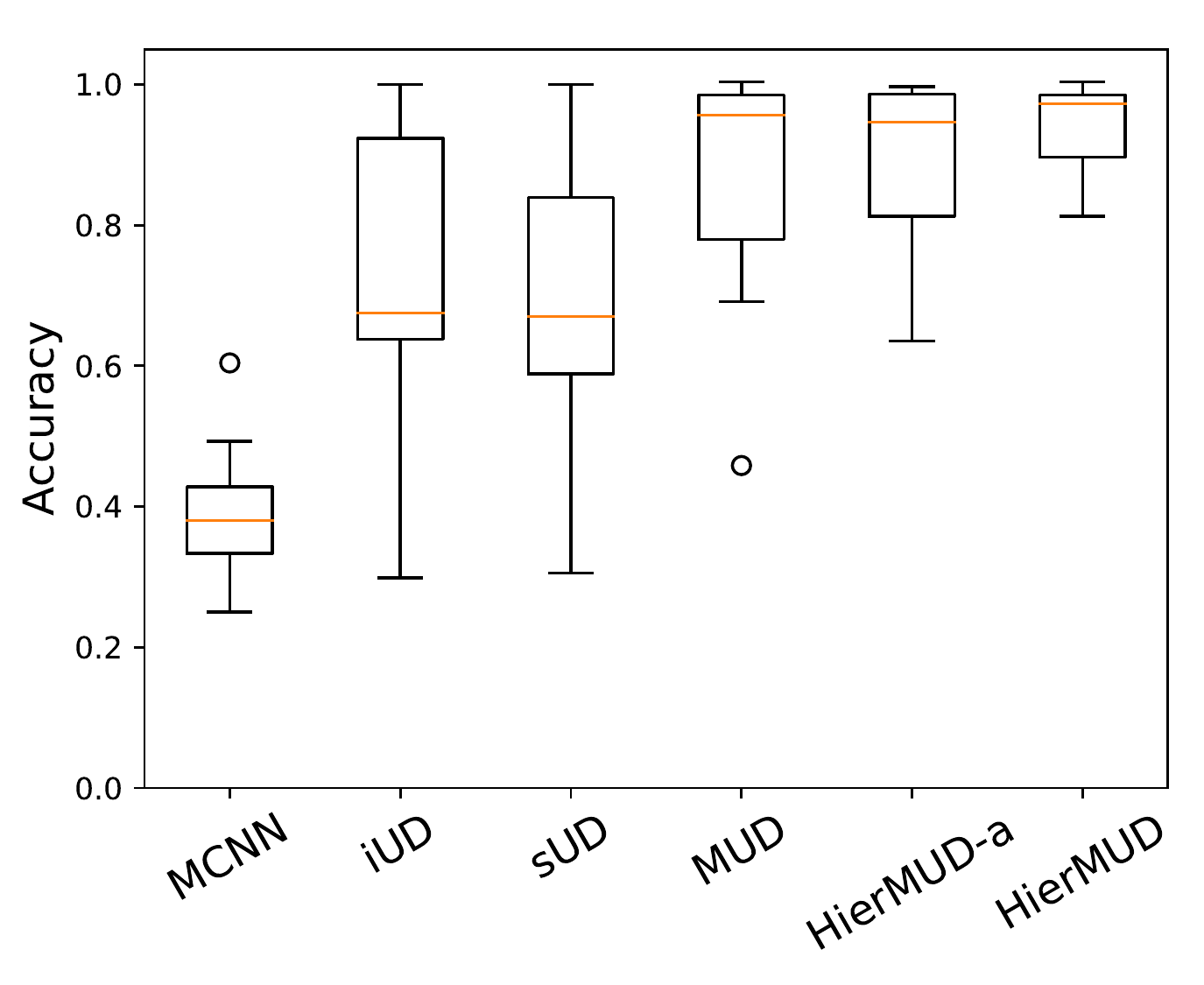}
        {(b) B1$\to$B2}
        \label{fig:lf}
    \end{minipage}
    \begin{minipage}{0.33\textwidth}
        \centering
        \includegraphics[width=1\linewidth]{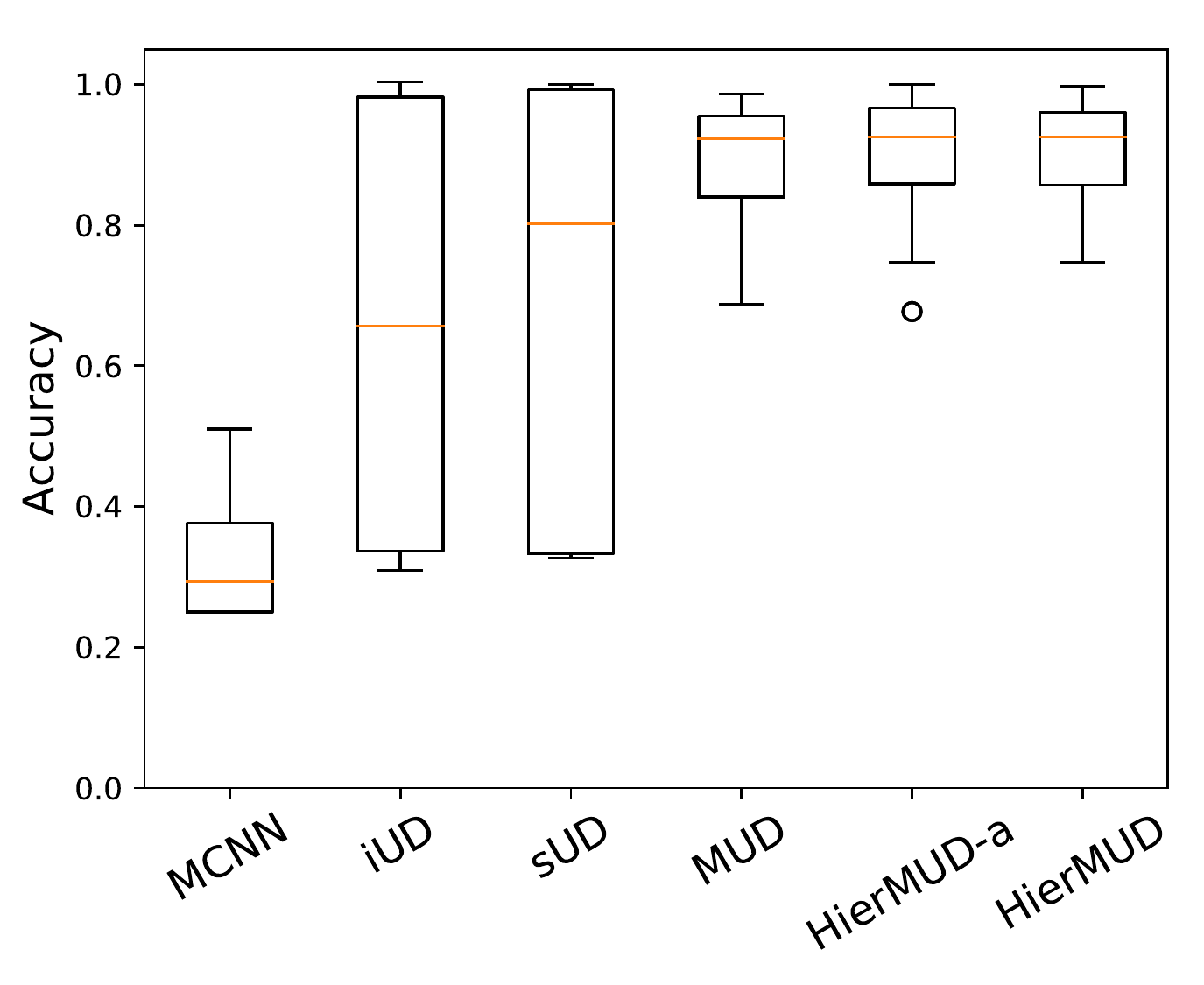}
        {(c) B2$\to$B1}
        \label{fig:lb}
    \end{minipage}
    \caption{The performance of baseline methods and our method for damage localization task on the lab-scale VBI dataset.}
    \label{fig:loc}
\end{figure*}

\begin{figure*}[htb]
    \centering
    \begin{minipage}{0.33\textwidth}
        \centering
        \includegraphics[width=1\linewidth]{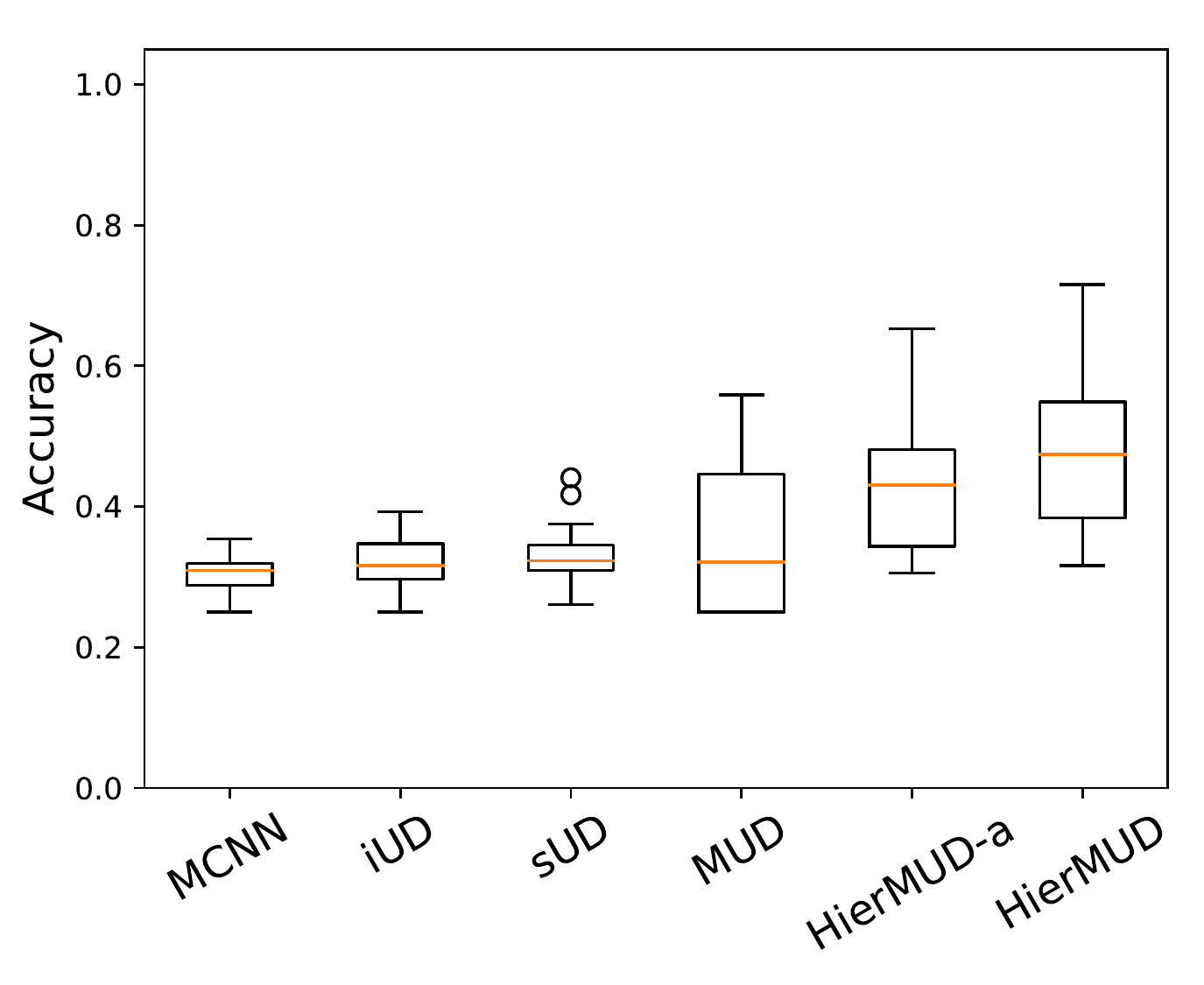}
        {(a) Overall}
        \label{fig:q}
    \end{minipage}%
    \begin{minipage}{0.33\textwidth}
        \centering
        \includegraphics[width=1\linewidth]{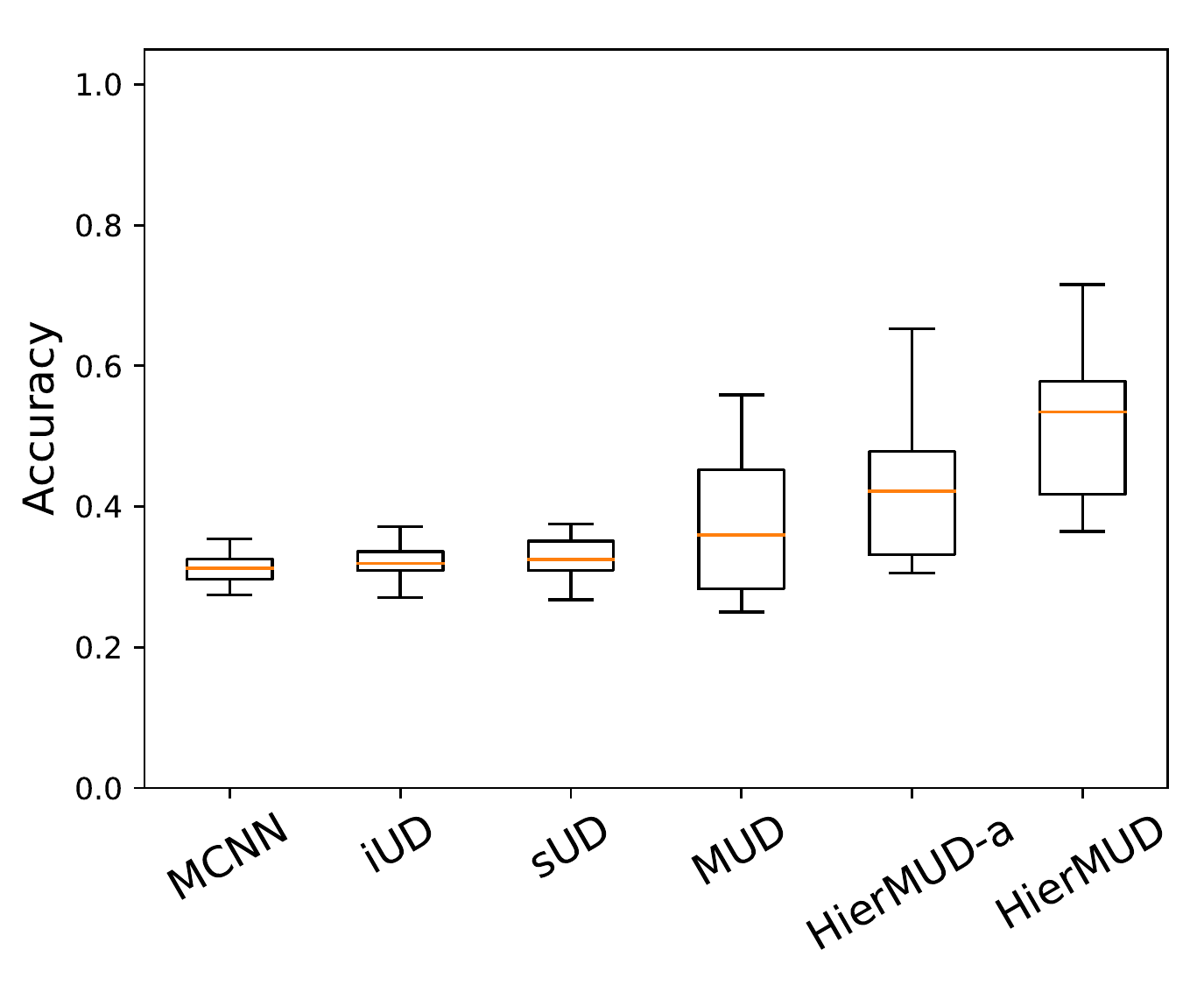}
        {(b) B1$\to$B2}
        \label{fig:qf}
    \end{minipage}
    \begin{minipage}{0.33\textwidth}
        \centering
        \includegraphics[width=1\linewidth]{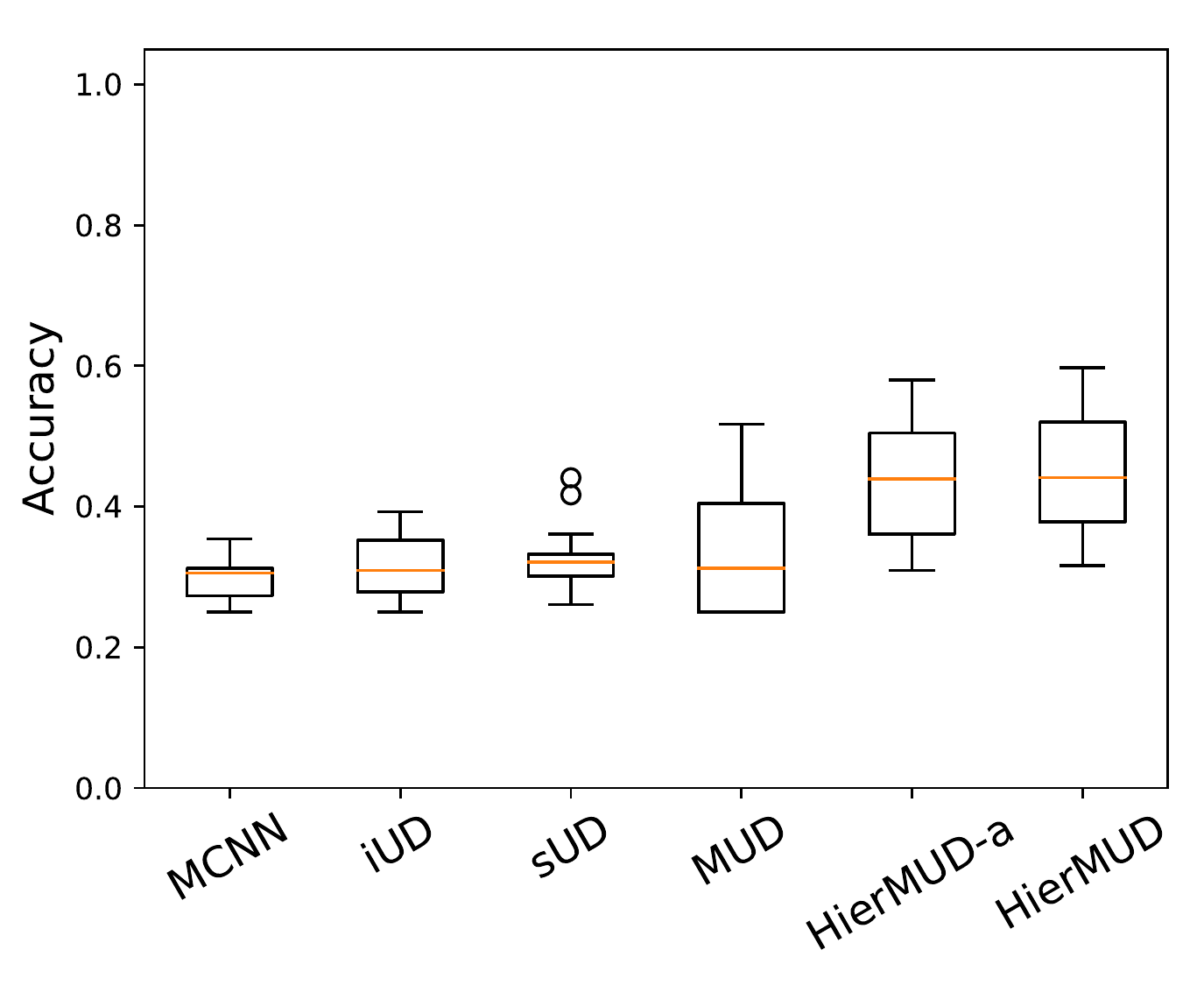}
        {(c) B2$\to$B1}
        \label{fig:qb}
    \end{minipage}
    \caption{The performance of baseline methods and our method for damage quantification task on the lab-scale VBI dataset.}
    \label{fig:qua}
\end{figure*}

Figures~\ref{fig:det},~\ref{fig:loc}, and~\ref{fig:qua} show the boxplots of performance metric in damage detection, localization, and quantification tasks, respectively. In each figure set, boxplot (a) shows the overall 60 test results of our method and baselines. Boxplots (b) and (c) present 30 test results for model transfer from B1 to B2 and from B2 to B1, respectively. Each box in the boxplot shows five values, including minimum, maximum, median, the first quartile, and the third quartile. In addition to the results from Table~\ref{tab:results}, we also observe that the results of our method have smaller variance (i.e., smaller box) than that of baseline methods, which indicates a more stable performance with different random initialization. Furthermore, for the damage quantification task using our method, results for model transfer from B1 to B2 are better than those for model transfer from B2 to B1. This difference is coincidental with or due to the fact that different model transfer directions have 
distinct model generalizability (i.e., the model learned from B1 data is more generalizable and easy to transfer than that learned from B2.). 

In summary, our method outperforms baselines in comprehensive laboratory evaluations, which shows the effectiveness of combining domain adversarial, multi-task learning with the hierarchical architecture and the soft-max objective.

\subsection{Characterizing the extracted feature distributions}
\begin{figure*}[htb]
    \centering
    \begin{minipage}{0.5\textwidth}
        \centering
        \includegraphics[width=1\linewidth]{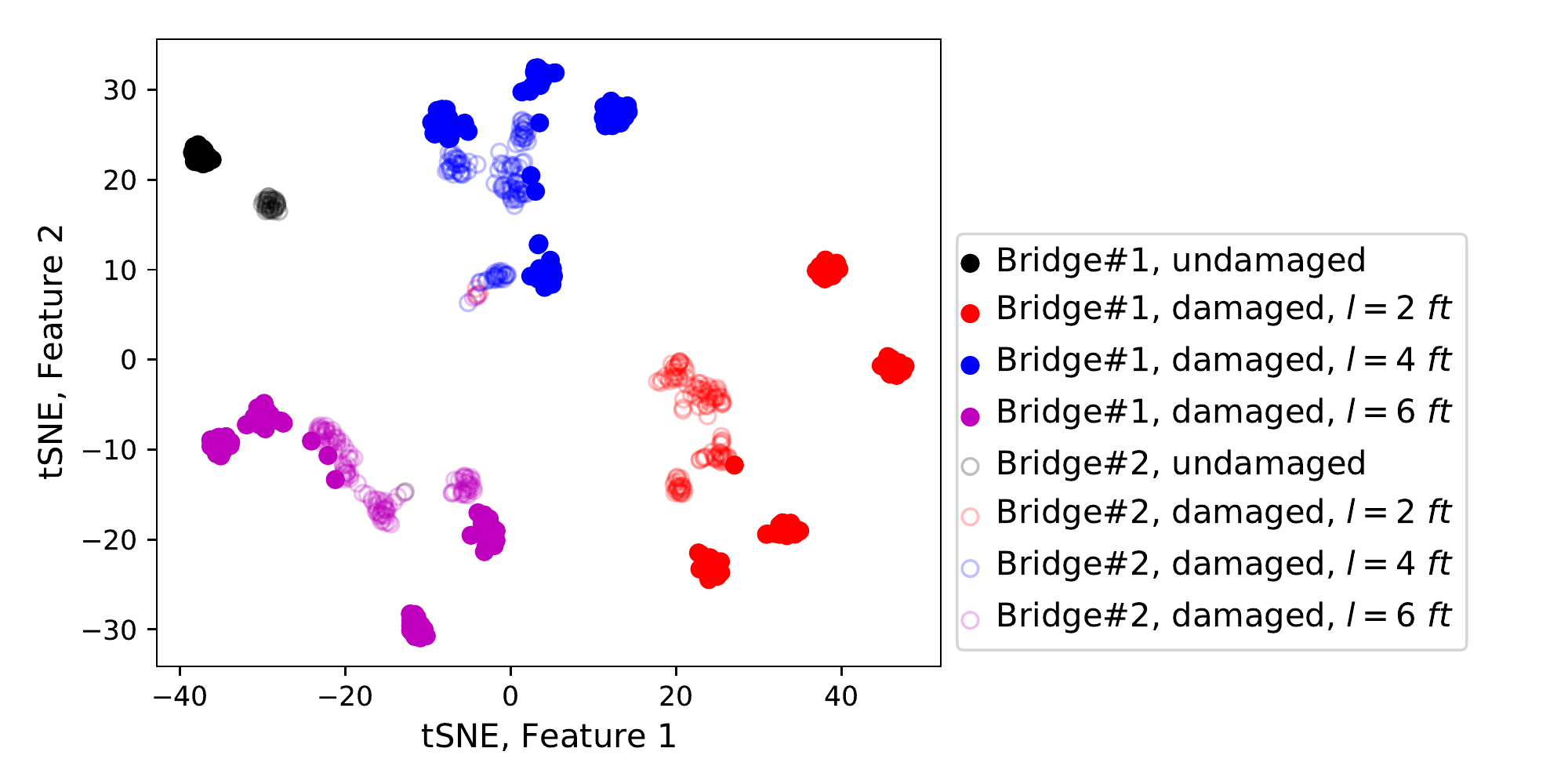}
        {(a)}
        \label{fig:ltsne}
    \end{minipage}%
    \begin{minipage}{0.5\textwidth}
        \centering
        \includegraphics[width=1\linewidth]{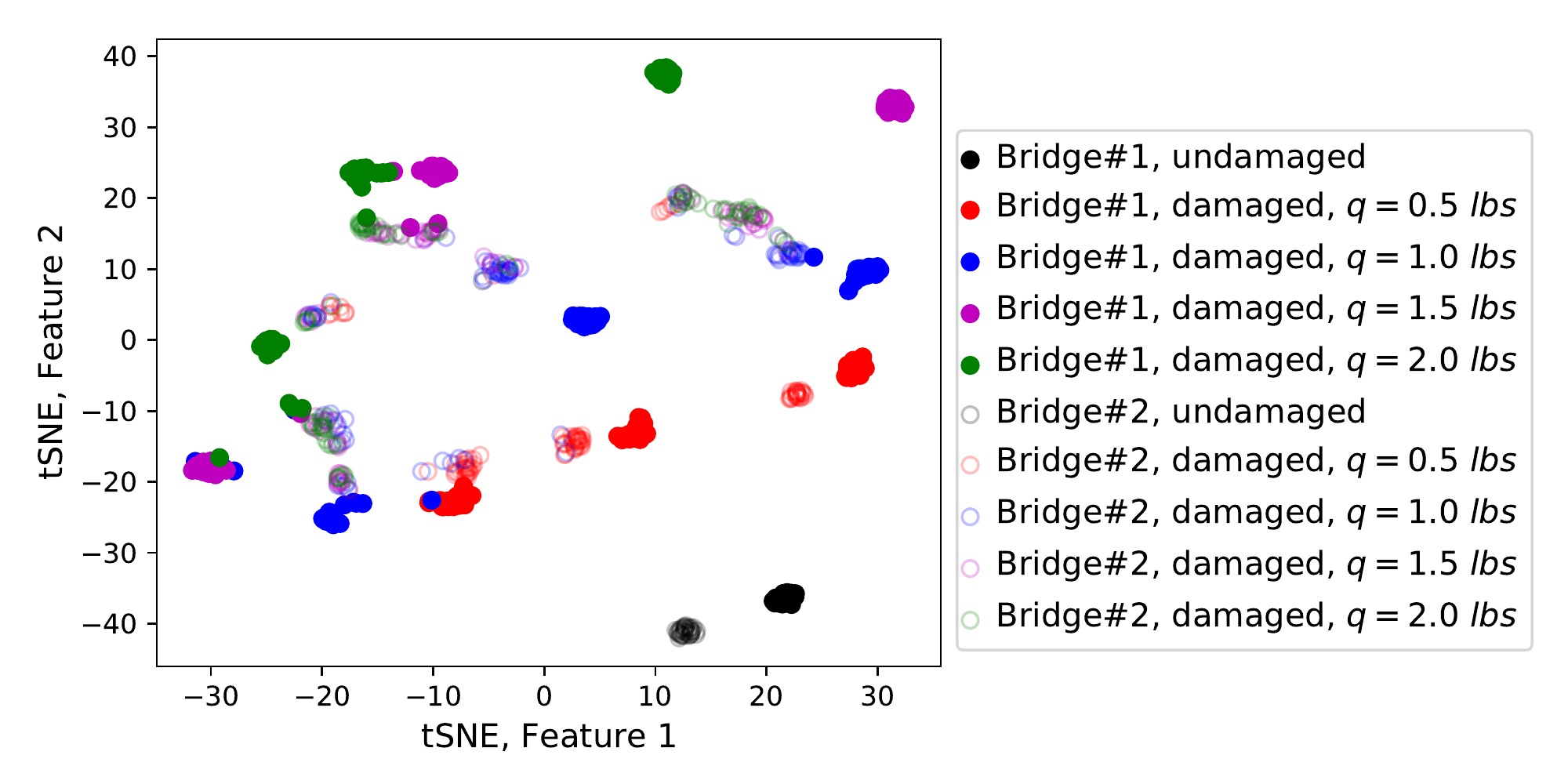}
        {(b)}
        \label{fig:q2tsne}
    \end{minipage}
    \caption{The 2D tSNE visualization of (a) the task-shared features after domain adaptation in the damage localization task and (b) the task-specific features after domain adaptation in the damage quantification task for the evaluation of model transfer from Bridge\#1 (B1) to Bridge\#2 (B2).}
    \label{fig:afterDA}
\end{figure*}

\begin{figure}[htb]
    \centering
    \includegraphics[width=1\linewidth]{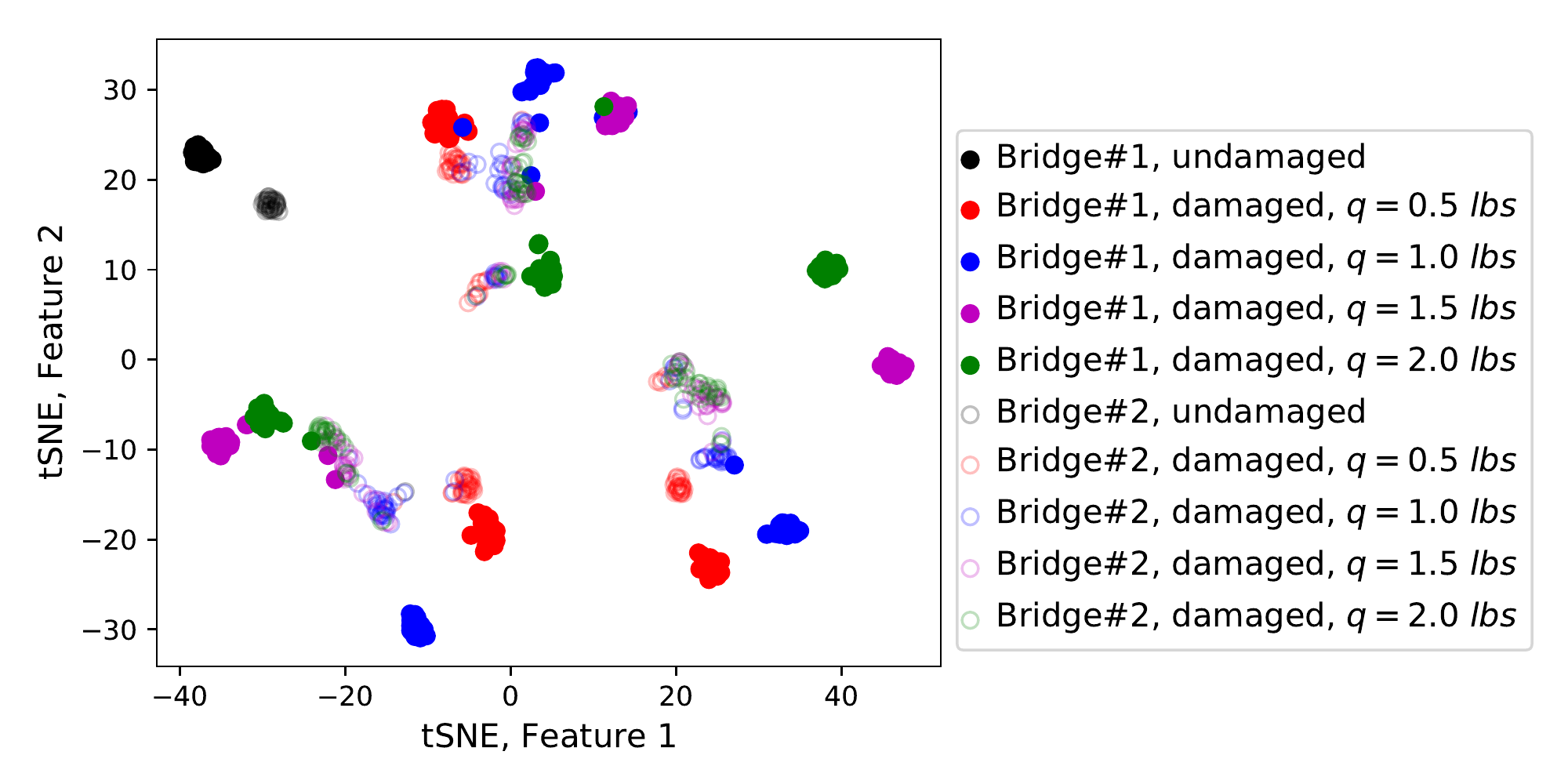}
    \caption{The 2D tSNE visualization of the task-shared features after domain adaptation in the damage quantification task for the evaluation of model transfer from Bridge\#1 (B1) to Bridge\#2 (B2).}
    \label{fig:qtsne}
\end{figure}

In this section, we characterize the extracted features to show the effect of domain adaptation on the distribution of the extracted features. Figure~\ref{fig:afterDA} and~\ref{fig:qtsne} shows the feature distributions plotted using tSNE to project feature distributions at different feature layers of the network into a two-dimensional feature space. Figure~\ref{fig:afterDA} (a) shows the tSNE projection of the task-shared features ($\bar{Z}_S$ and $\bar{Z}_T$) after domain adaptation for the damage localization task; Figure~\ref{fig:afterDA} (b) shows the tSNE project of the source and target domains' task-specific features (${Z}^h_S$ and ${Z}^h_T$) after domain adaptation for the damage quantification task. Different colors indicate different damage labels. Filled and unfilled markers represent features of source and target domains, respectively. Comparing the distributions of features before and after domain adaptation in Figure~\ref{fig:beforDA} and~\ref{fig:afterDA}, we observe that our domain adaptation method successfully transformed the feature distributions of the two domains to be much more similar to each other. Also, comparing Figure~\ref{fig:afterDA} (a) and (b), the distributions of the features are more similar and thus better matched for the easy-to-learn damage localization task than that for the hard-to-learn damage quantification task.

Further, Figure~\ref{fig:qtsne} shows the same tSNE projection of the task-shared features ($\bar{Z}_S$ and $\bar{Z}_T$) but color-coded with different damage severity levels. The distributions of task-shared features having the same damage severity label in Figure~\ref{fig:qtsne} (i.e., the markers with the same color) are more spread out than those in Figure~\ref{fig:afterDA} (b), making features in Figure~\ref{fig:qtsne} more difficult to be correctly classified. Therefore, task-specific features are more task-informative than task-shared features for the quantification task, which corresponds to our results that using the task-specific hierarchical features outperforms using the task-shared features for damage quantification. 

\subsection{Characterizing training process}
\begin{figure}[htb]
    \centering
    \includegraphics[width=1\linewidth]{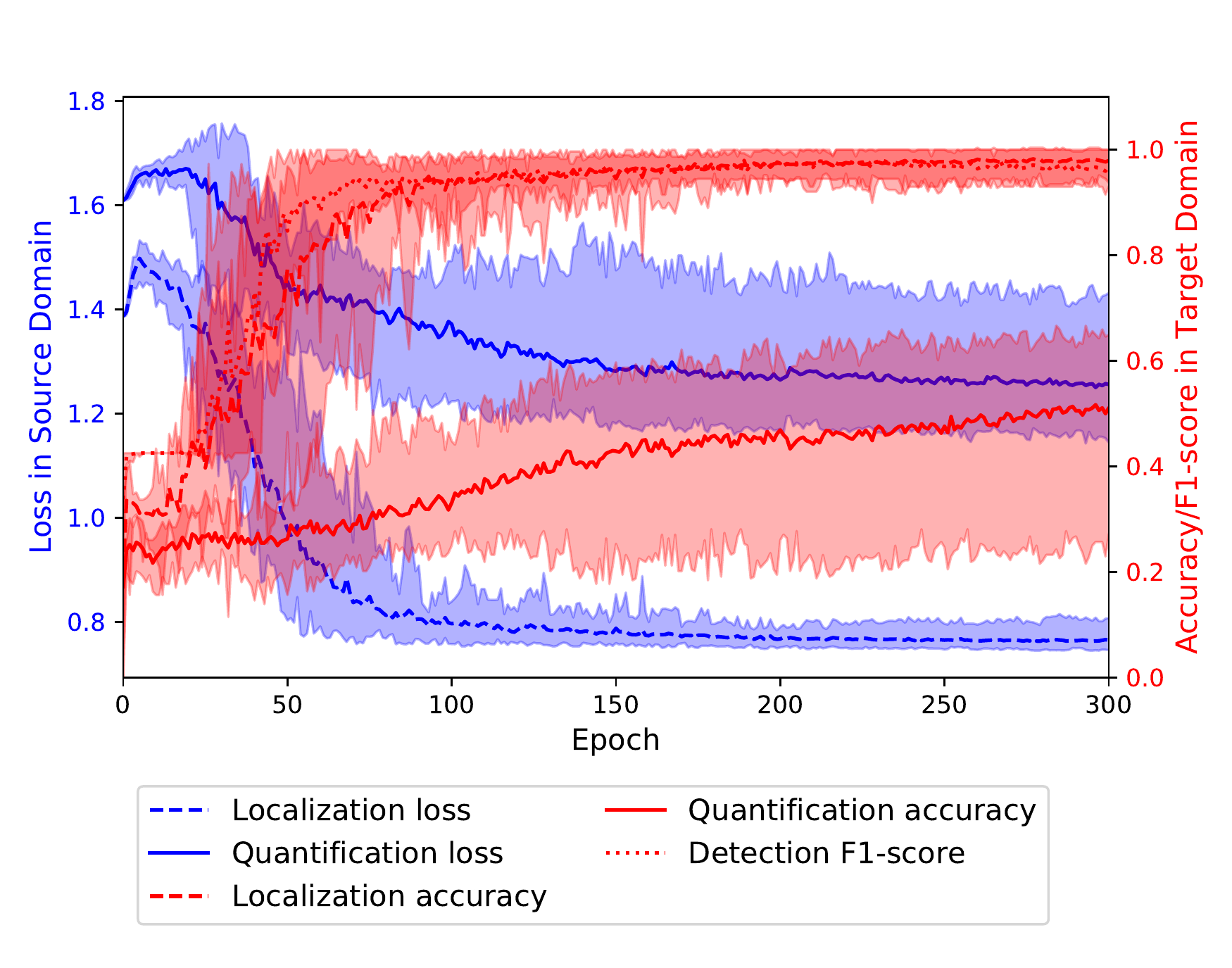}
    \caption{The losses of damage localization and quantification in the source domain, and the accuracy/F1-score of damage detection, localization, and quantification in the target domain.}
    \label{fig:loss}
\end{figure}

To characterize the training process, we show the training losses and testing accuracy for model transfer from B1 to B2 using V2 data in Figure~\ref{fig:loss}. Blue curves are average losses of the two task predictors in the source domain, and red curves are average accuracy/F1-score of damage detection, localization, and quantification in the target domain. The envelope covers loss or accuracy for the ten tests with different random initialization. The figure shows that the two losses fluctuate and even increase during early training steps (i.e., when the number of epochs is small) due to domain adversarial learning. They converge to small loss values after around 200 epochs. We also observe that the quantification loss and accuracy have a large area of the envelope because it is difficult to find optimized saddle points of model parameters for this hard-to-learn quantification task.

\section{Conclusion and Future Work}
In this work, we introduce HierMUD, a new multi-task unsupervised domain adaptation framework that uses drive-by vehicle acceleration responses to diagnose structural damage of multiple bridges without requiring labeled data from every bridge. Our framework jointly optimizes hierarchical feature extractors, damage predictors, and domain classifiers in an adversarial way to extract features that are task-informative and domain-invariant. To match distinct data distributions over multiple tasks, we introduce a novel loss function based on a new provable generalization risk bound to adaptively provide a larger gradient magnitude on matching tasks having more shifted distribution between the source and target domains. To learn multiple tasks with varying task difficulties, the feature extractor is designed to formulate a feature hierarchy, which learns two-level features: task-shared features are used for learning easy-to-learn tasks, and task-specific features are further extracted from task-shared features to learn hard-to-learn tasks. To the best of our knowledge, this is the first framework to transfer the model learned from one bridge to detect, localize, and quantify damage to another (target) bridge without any labels of the target bridge. We evaluate our framework on experimental data collected from three vehicles passing over two bridges individually. Our framework consistently outperforms baseline methods and has smaller prediction variance. It achieves average accuracy values of 95\% for damage detection, 93\% for damage localization, and up to 48\% for damage quantification.

In future work, we plan to investigate how the model transfer direction affects the domain adaptation performance. For instance, we will investigate what characteristics of source domain bridges lead to a better performance when the model is transferred to a target domain. In addition, we intend to study how to transfer our multi-task damage diagnosis model between bridges with different structural forms (e.g., truss bridge, suspension bridge, etc.). Moreover, we believe this framework can be generalized to other multi-task UDA problems beyond structural health monitoring. Thus, we plan to apply our framework in other applications (e.g., visual recognition, machine monitoring, etc.) to further  evaluate its robustness and effectiveness.

\begin{acks}
This research was supported in part by the Leavell Fellowship on Sustainable Built Environment and UPS endowment fund at Stanford University, and Mobility 21 at Carnegie Mellon University.
\end{acks}

\bibliographystyle{SageH}
\bibliography{citation.bib}

\begin{thebibliography}{56}
\providecommand{\natexlab}[1]{#1}
\providecommand{\url}[1]{\texttt{#1}}
\providecommand{\urlprefix}{URL }
\expandafter\ifx\csname urlstyle\endcsname\relax
  \providecommand{\doi}[1]{DOI:\discretionary{}{}{}#1}\else
  \providecommand{\doi}{DOI:\discretionary{}{}{}\begingroup
  \urlstyle{rm}\Url}\fi

\bibitem[{ASCE(2017)}]{asce}
ASCE (2017) Bridges-infrastructure report card.
\newblock
  \url{https://www.infrastructurereportcard.org/wp-content/uploads/2017/01/Bridges-Final.pdf}.

\bibitem[{Augenstein et~al.(2018)Augenstein, Ruder and
  S{\o}gaard}]{augenstein2018multi}
Augenstein I, Ruder S and S{\o}gaard A (2018) Multi-task learning of pairwise
  sequence classification tasks over disparate label spaces.
\newblock \emph{arXiv preprint arXiv:1802.09913} .

\bibitem[{Ben-David et~al.(2010)Ben-David, Blitzer, Crammer, Kulesza, Pereira
  and Vaughan}]{ben2010theory}
Ben-David S, Blitzer J, Crammer K, Kulesza A, Pereira F and Vaughan JW (2010) A
  theory of learning from different domains.
\newblock \emph{Machine learning} 79(1-2): 151--175.

\bibitem[{Cao et~al.(2018)Cao, Long and Wang}]{cao2018unsupervised}
Cao Y, Long M and Wang J (2018) Unsupervised domain adaptation with
  distribution matching machines.
\newblock In: \emph{Proceedings of the AAAI Conference on Artificial
  Intelligence}, volume~32.

\bibitem[{Caruana(1997)}]{caruana1997multitask}
Caruana R (1997) Multitask learning.
\newblock \emph{Machine learning} 28(1): 41--75.

\bibitem[{Cerda et~al.(2014)Cerda, Chen, Bielak, Garrett, Rizzo and
  Kovacevic}]{cerda2014indirect}
Cerda F, Chen S, Bielak J, Garrett JH, Rizzo P and Kovacevic J (2014) Indirect
  structural health monitoring of a simplified laboratory-scale bridge model.
\newblock \emph{Smart Structures and Systems} 13(5): 849--868.

\bibitem[{Deraemaeker and Worden(2018)}]{DERAEMAEKER20181}
Deraemaeker A and Worden K (2018) A comparison of linear approaches to filter
  out environmental effects in structural health monitoring.
\newblock \emph{Mechanical Systems and Signal Processing} 105: 1 -- 15.

\bibitem[{Dong et~al.(2015)Dong, Wu, He, Yu and Wang}]{dong2015multi}
Dong D, Wu H, He W, Yu D and Wang H (2015) Multi-task learning for multiple
  language translation.
\newblock In: \emph{Proceedings of the 53rd Annual Meeting of the Association
  for Computational Linguistics and the 7th International Joint Conference on
  Natural Language Processing (Volume 1: Long Papers)}. pp. 1723--1732.

\bibitem[{Eshkevari and Pakzad(2020)}]{eshkevari2020signal}
Eshkevari SS and Pakzad SN (2020) Signal reconstruction from mobile sensors
  network using matrix completion approach.
\newblock In: \emph{Topics in Modal Analysis \& Testing, Volume 8}. Springer,
  pp. 61--75.

\bibitem[{Ganin et~al.(2016)Ganin, Ustinova, Ajakan, Germain, Larochelle,
  Laviolette, Marchand and Lempitsky}]{ganin2016domain}
Ganin Y, Ustinova E, Ajakan H, Germain P, Larochelle H, Laviolette F, Marchand
  M and Lempitsky V (2016) Domain-adversarial training of neural networks.
\newblock \emph{The Journal of Machine Learning Research} 17(1): 2096--2030.

\bibitem[{Gonz{\'a}lez et~al.(2012)Gonz{\'a}lez, OBrien and
  McGetrick}]{gonzalez2012identification}
Gonz{\'a}lez A, OBrien EJ and McGetrick P (2012) Identification of damping in a
  bridge using a moving instrumented vehicle.
\newblock \emph{Journal of Sound and Vibration} 331(18): 4115--4131.

\bibitem[{Goodfellow et~al.(2016)Goodfellow, Bengio, Courville and
  Bengio}]{goodfellow2016deep}
Goodfellow I, Bengio Y, Courville A and Bengio Y (2016) \emph{Deep learning},
  volume~1.
\newblock MIT press Cambridge.

\bibitem[{Guo et~al.(2018)Guo, Haque, Huang, Yeung and
  Fei-Fei}]{guo2018dynamic}
Guo M, Haque A, Huang DA, Yeung S and Fei-Fei L (2018) Dynamic task
  prioritization for multitask learning.
\newblock In: \emph{Proceedings of the European Conference on Computer Vision
  (ECCV)}. pp. 270--287.

\bibitem[{Hartle et~al.(2002)Hartle, Ryan, Mann, Danovich, Sosko, Bouscher,
  Baker~Jr et~al.}]{hartle2002bridge}
Hartle RA, Ryan TW, Mann E, Danovich LJ, Sosko WB, Bouscher JW, Baker~Jr M
  et~al. (2002) Bridge inspector's reference manual: Volume 1 and volume 2.
\newblock Technical report, United States. Federal Highway Administration.

\bibitem[{Hashimoto et~al.(2016)Hashimoto, Xiong, Tsuruoka and
  Socher}]{hashimoto2016joint}
Hashimoto K, Xiong C, Tsuruoka Y and Socher R (2016) A joint many-task model:
  Growing a neural network for multiple nlp tasks.
\newblock \emph{arXiv preprint arXiv:1611.01587} .

\bibitem[{Jakubovitz et~al.(2019)Jakubovitz, Giryes and
  Rodrigues}]{jakubovitz2019generalization}
Jakubovitz D, Giryes R and Rodrigues MR (2019) Generalization error in deep
  learning.
\newblock In: \emph{Compressed Sensing and Its Applications}. Springer, pp.
  153--193.

\bibitem[{Jiang and Zhai(2007)}]{jiang2007instance}
Jiang J and Zhai C (2007) Instance weighting for domain adaptation in nlp.
\newblock ACL.

\bibitem[{Jou and Chang(2016)}]{jou2016deep}
Jou B and Chang SF (2016) Deep cross residual learning for multitask visual
  recognition.
\newblock In: \emph{Proceedings of the 24th ACM international conference on
  Multimedia}. pp. 998--1007.

\bibitem[{Kenny et~al.(2012)Kenny, Fluck and Jetson}]{kenny2012placing}
Kenny J, Fluck A and Jetson T (2012) Placing a value on academic work: The
  development and implementation of a time-based academic workload model.
\newblock \emph{Australian Universities' Review, The} 54(2): 50--60.

\bibitem[{Kenny and Fluck(2014)}]{kenny2014effectiveness}
Kenny JD and Fluck AE (2014) The effectiveness of academic workload models in
  an institution: a staff perspective.
\newblock \emph{Journal of Higher Education Policy and Management} 36(6):
  585--602.

\bibitem[{Lederman et~al.(2014)Lederman, Wang, Bielak, Noh, Garrett, Chen,
  Kovacevic, Cerda and Rizzo}]{lederman2014damage}
Lederman G, Wang Z, Bielak J, Noh H, Garrett JH, Chen S, Kovacevic J, Cerda F
  and Rizzo P (2014) Damage quantification and localization algorithms for
  indirect shm of bridges.
\newblock In: \emph{Proc. Int. Conf. Bridge Maint., Safety Manag., Shanghai,
  China}.

\bibitem[{Lin and Yang(2005)}]{lin2005use}
Lin C and Yang Y (2005) Use of a passing vehicle to scan the fundamental bridge
  frequencies: An experimental verification.
\newblock \emph{Engineering Structures} 27(13): 1865--1878.

\bibitem[{Liu(2020)}]{liu2020scalable}
Liu J (2020) Scalable bridge health monitoring using drive-by vehicles: Phd
  forum abstract.
\newblock In: \emph{Proceedings of the 18th Conference on Embedded Networked
  Sensor Systems}. pp. 817--818.

\bibitem[{Liu et~al.(2019{\natexlab{a}})Liu, Berg{\'e}s, Bielak, Garrett,
  Kova{\v{c}}evi{\'c} and Noh}]{liu2019damage}
Liu J, Berg{\'e}s M, Bielak J, Garrett JH, Kova{\v{c}}evi{\'c} J and Noh HY
  (2019{\natexlab{a}}) A damage localization and quantification algorithm for
  indirect structural health monitoring of bridges using multi-task learning.
\newblock In: \emph{AIP Conference Proceedings}, volume 2102. AIP Publishing
  LLC, p. 090003.

\bibitem[{Liu et~al.(2020{\natexlab{a}})Liu, Chen, Chen, Berg{\'e}s, Bielak and
  Noh}]{liu2020damage}
Liu J, Chen B, Chen S, Berg{\'e}s M, Bielak J and Noh H (2020{\natexlab{a}})
  Damage-sensitive and domain-invariant feature extraction for
  vehicle-vibration-based bridge health monitoring.
\newblock In: \emph{ICASSP 2020-2020 IEEE International Conference on
  Acoustics, Speech and Signal Processing (ICASSP)}. IEEE, pp. 3007--3011.

\bibitem[{Liu et~al.(2020{\natexlab{b}})Liu, Chen, Berg{\'e}s, Bielak, Garrett,
  Kova{\v{c}}evi{\'c} and Noh}]{liu2020diagnosis}
Liu J, Chen S, Berg{\'e}s M, Bielak J, Garrett JH, Kova{\v{c}}evi{\'c} J and
  Noh HY (2020{\natexlab{b}}) Diagnosis algorithms for indirect structural
  health monitoring of a bridge model via dimensionality reduction.
\newblock \emph{Mechanical Systems and Signal Processing} 136: 106454.

\bibitem[{Liu et~al.(2019{\natexlab{b}})Liu, Xu, Berg{\'e}s, Bielak, GARRETT
  and Noh}]{liu2019expectation}
Liu J, Xu S, Berg{\'e}s M, Bielak J, GARRETT JH and Noh HY (2019{\natexlab{b}})
  An expectation-maximization algorithm-based framework for
  vehicle-vibration-based indirect structural health monitoring of bridges.
\newblock \emph{Structural Health Monitoring 2019} .

\bibitem[{Long et~al.(2015)Long, Cao, Wang and Jordan}]{long2015learning}
Long M, Cao Y, Wang J and Jordan M (2015) Learning transferable features with
  deep adaptation networks.
\newblock In: \emph{International conference on machine learning}. PMLR, pp.
  97--105.

\bibitem[{Luo et~al.(2020)Luo, Ren, Dao-Qing and Yan}]{luo2020unsupervised}
Luo YW, Ren CX, Dao-Qing D and Yan H (2020) Unsupervised domain adaptation via
  discriminative manifold propagation.
\newblock \emph{IEEE Transactions on Pattern Analysis and Machine Intelligence}
  .

\bibitem[{Luong et~al.(2015)Luong, Le, Sutskever, Vinyals and
  Kaiser}]{luong2015multi}
Luong MT, Le QV, Sutskever I, Vinyals O and Kaiser L (2015) Multi-task sequence
  to sequence learning.
\newblock \emph{arXiv preprint arXiv:1511.06114} .

\bibitem[{Malekjafarian et~al.(2015)Malekjafarian, McGetrick and
  OBrien}]{malekjafarian2015review}
Malekjafarian A, McGetrick PJ and OBrien EJ (2015) A review of indirect bridge
  monitoring using passing vehicles.
\newblock \emph{Shock and vibration} 2015.

\bibitem[{Malekjafarian and OBrien(2014)}]{malekjafarian2014identification}
Malekjafarian A and OBrien EJ (2014) Identification of bridge mode shapes using
  short time frequency domain decomposition of the responses measured in a
  passing vehicle.
\newblock \emph{Engineering Structures} 81: 386--397.

\bibitem[{Maurer et~al.(2016)Maurer, Pontil and
  Romera-Paredes}]{maurer2016benefit}
Maurer A, Pontil M and Romera-Paredes B (2016) The benefit of multitask
  representation learning.
\newblock \emph{The Journal of Machine Learning Research} 17(1): 2853--2884.

\bibitem[{McGetrick and Kim(2013)}]{mcgetrick2013parametric}
McGetrick PJ and Kim CW (2013) A parametric study of a drive by bridge
  inspection system based on the morlet wavelet.
\newblock In: \emph{Key Engineering Materials}, volume 569. Trans Tech Publ,
  pp. 262--269.

\bibitem[{Mei et~al.(2019)Mei, G{\"u}l and Boay}]{mei2019indirect}
Mei Q, G{\"u}l M and Boay M (2019) Indirect health monitoring of bridges using
  mel-frequency cepstral coefficients and principal component analysis.
\newblock \emph{Mechanical Systems and Signal Processing} 119: 523--546.

\bibitem[{Misra et~al.(2016)Misra, Shrivastava, Gupta and
  Hebert}]{misra2016cross}
Misra I, Shrivastava A, Gupta A and Hebert M (2016) Cross-stitch networks for
  multi-task learning.
\newblock In: \emph{Proceedings of the IEEE conference on computer vision and
  pattern recognition}. pp. 3994--4003.

\bibitem[{Nasrollahi et~al.(2018)Nasrollahi, Deng, Ma and
  Rizzo}]{doi:10.1177/1475921717699375}
Nasrollahi A, Deng W, Ma Z and Rizzo P (2018) Multimodal structural health
  monitoring based on active and passive sensing.
\newblock \emph{Structural Health Monitoring} 17(2): 395--409.

\bibitem[{Nguyen and Tran(2010)}]{nguyen2010multi}
Nguyen KV and Tran HT (2010) Multi-cracks detection of a beam-like structure
  based on the on-vehicle vibration signal and wavelet analysis.
\newblock \emph{Journal of Sound and Vibration} 329(21): 4455--4465.

\bibitem[{Pan et~al.(2010)Pan, Tsang, Kwok and Yang}]{pan2010domain}
Pan SJ, Tsang IW, Kwok JT and Yang Q (2010) Domain adaptation via transfer
  component analysis.
\newblock \emph{IEEE Transactions on Neural Networks} 22(2): 199--210.

\bibitem[{Paszke et~al.(2017)Paszke, Gross, Chintala, Chanan, Yang, DeVito,
  Lin, Desmaison, Antiga and Lerer}]{paszke2017automatic}
Paszke A, Gross S, Chintala S, Chanan G, Yang E, DeVito Z, Lin Z, Desmaison A,
  Antiga L and Lerer A (2017) Automatic differentiation in pytorch .

\bibitem[{Sadeghi~Eshkevari et~al.(2020)Sadeghi~Eshkevari, Pakzad,
  Tak{\'a}{\v{c}} and Matarazzo}]{sadeghi2020modal}
Sadeghi~Eshkevari S, Pakzad SN, Tak{\'a}{\v{c}} M and Matarazzo TJ (2020) Modal
  identification of bridges using mobile sensors with sparse vibration data.
\newblock \emph{Journal of Engineering Mechanics} 146(4): 04020011.

\bibitem[{Saito et~al.(2018)Saito, Watanabe, Ushiku and
  Harada}]{saito2018maximum}
Saito K, Watanabe K, Ushiku Y and Harada T (2018) Maximum classifier
  discrepancy for unsupervised domain adaptation.
\newblock In: \emph{Proceedings of the IEEE conference on computer vision and
  pattern recognition}. pp. 3723--3732.

\bibitem[{Sun et~al.(2020)Sun, Shang, Xia, Bhowmick and
  Nagarajaiah}]{sun2020review}
Sun L, Shang Z, Xia Y, Bhowmick S and Nagarajaiah S (2020) Review of bridge
  structural health monitoring aided by big data and artificial intelligence:
  from condition assessment to damage detection.
\newblock \emph{Journal of Structural Engineering} 146(5): 04020073.

\bibitem[{Taddei et~al.(2018)Taddei, Penn, Yano and Patera}]{Taddei2018}
Taddei T, Penn JD, Yano M and Patera AT (2018) Simulation-based classification;
  a model-order-reduction approach for structural health monitoring.
\newblock \emph{Archives of Computational Methods in Engineering} 25(1):
  23--45.

\bibitem[{Van~der Maaten and Hinton(2008)}]{van2008visualizing}
Van~der Maaten L and Hinton G (2008) Visualizing data using t-sne.
\newblock \emph{Journal of machine learning research} 9(11).

\bibitem[{Wan and Ni(2019)}]{wan2019bayesian}
Wan HP and Ni YQ (2019) Bayesian multi-task learning methodology for
  reconstruction of structural health monitoring data.
\newblock \emph{Structural Health Monitoring} 18(4): 1282--1309.

\bibitem[{Xu and Noh(2021)}]{xu2021phymdan}
Xu S and Noh HY (2021) Phymdan: Physics-informed knowledge transfer between
  buildings for seismic damage diagnosis through adversarial learning.
\newblock \emph{Mechanical Systems and Signal Processing} 151: 107374.

\bibitem[{Yang and Hospedales(2016)}]{yang2016deep}
Yang Y and Hospedales T (2016) Deep multi-task representation learning: A
  tensor factorisation approach.
\newblock \emph{arXiv preprint arXiv:1605.06391} .

\bibitem[{Yang et~al.(2014)Yang, Li and Chang}]{yang2014constructing}
Yang Y, Li Y and Chang KC (2014) Constructing the mode shapes of a bridge from
  a passing vehicle: a theoretical study.
\newblock \emph{Smart Structures and Systems} 13(5): 797--819.

\bibitem[{Yang et~al.(2020{\natexlab{a}})Yang, Wang, Shi, Xu and
  Wu}]{yang2020state}
Yang Y, Wang ZL, Shi K, Xu H and Wu Y (2020{\natexlab{a}}) State-of-the-art of
  the vehicle-based methods for detecting the various properties of highway
  bridges and railway tracks.
\newblock \emph{International Journal of Structural Stability and Dynamics} :
  2041004.

\bibitem[{Yang et~al.(2004)Yang, Lin and Yau}]{yang2004extracting}
Yang YB, Lin C and Yau J (2004) Extracting bridge frequencies from the dynamic
  response of a passing vehicle.
\newblock \emph{Journal of Sound and Vibration} 272(3-5): 471--493.

\bibitem[{Yang et~al.(2020{\natexlab{b}})Yang, Yang, Zhang and
  Wu}]{yang2020vehicle}
Yang YB, Yang JP, Zhang B and Wu Y (2020{\natexlab{b}}) \emph{Vehicle Scanning
  Method for Bridges}.
\newblock Wiley Online Library.

\bibitem[{Zhang and Gao(2019)}]{zhang2019transfer}
Zhang L and Gao X (2019) Transfer adaptation learning: A decade survey.
\newblock \emph{arXiv preprint arXiv:1903.04687} .

\bibitem[{Zhang et~al.(2019)Zhang, Liu, Long and Jordan}]{zhang2019bridging}
Zhang Y, Liu T, Long M and Jordan MI (2019) Bridging theory and algorithm for
  domain adaptation.
\newblock \emph{arXiv preprint arXiv:1904.05801} .

\bibitem[{Zhao et~al.(2018)Zhao, Zhang, Wu, Moura, Costeira and
  Gordon}]{zhao2018adversarial}
Zhao H, Zhang S, Wu G, Moura JM, Costeira JP and Gordon GJ (2018) Adversarial
  multiple source domain adaptation.
\newblock In: \emph{Advances in neural information processing systems}. pp.
  8559--8570.

\bibitem[{Zhong et~al.(2010)Zhong, Fan, Yang, Verscheure and
  Ren}]{zhong2010cross}
Zhong E, Fan W, Yang Q, Verscheure O and Ren J (2010) Cross validation
  framework to choose amongst models and datasets for transfer learning.
\newblock In: \emph{Joint European Conference on Machine Learning and Knowledge
  Discovery in Databases}. Springer, pp. 547--562.

\end{thebibliography}

\newpage
\section{Appendix}
\begin{theorem}
Let $\mathcal{W}$ be a hypothesis space on $\mathcal{X}$ with VC dimension $d_W$ and $\mathcal{H}$ be a hypothesis space on $\mathcal{Z}$ with VC dimension $d_H$. If ${X}_S$ and ${X}_T$ are samples of size $N$ from $\mathcal{D}_S^X$ and $\mathcal{D}_T^X$, respectively, and ${Z}_S$ and ${Z}_T$ follow distributions $\mathcal{D}_S^Z$ and $\mathcal{D}_T^Z$, respectively,, then for any $\delta\in(0,1)$ with probability at least $1-\delta$, for every $h\in\mathcal{H}$ and $w\in\mathcal{W}$:
\begin{align*}
    \epsilon_T(h\circ w;f_T)
    \leq& \epsilon_S(h\circ w;f_S)+2\epsilon_S(h\circ w^*,f_S)\\&+\frac{1}{2}d_{\mathcal{H}\Delta\mathcal{H}}({Z}_T,{Z}_S)\\
    &+\mathcal{O}\Big(\sqrt{\frac{2d_H\log{(2N)+\log(2/\delta)}}{N}}\Big)\\
    &+\frac{1}{2}\sup_{\hat{h}\in\mathcal{H}}\big[d_{\hat{h},\mathcal{W}\Delta\mathcal{W}}({X}_S,{X}_T)\big]\\
    &+\mathcal{O}\Big(\sqrt{\frac{2d_W\log{(2N)+\log(2/\delta)}}{N}}\Big)\\
    &+\epsilon_T(h^*\circ w^*;f_T)+\epsilon_S(h^*\circ w^*;f_S),
\end{align*}

\begin{proof}
This proof relies on the triangle inequality for classification error.
\begingroup
\allowdisplaybreaks
    \begin{align*}
    \epsilon_T&(h\circ w;f_T)\\
    \leq& \epsilon_T(h\circ w;h^*\circ w^*)+\epsilon_T(h^*\circ w^*;f_T)\\
    \leq& \epsilon_T(h\circ w;h\circ w^*)+\epsilon_T(h\circ w^*;h^*\circ w^*)\\
    &+\epsilon_T(h^*\circ w^*;f_T)\\
    \leq& \epsilon_T(h^*\circ w^*;f_T)+\epsilon_T(h\circ w;h\circ w^*)\\
    &+\epsilon_S(h\circ w^*;h^*\circ w^*)\\
    &+|\epsilon_T(h\circ w^*;h^*\circ w^*)-\epsilon_S(h\circ w^*;h^*\circ w^*)|\\
    \leq& \epsilon_T(h^*\circ w^*;f_T)+\epsilon_T(h\circ w;h\circ w^*)\\
    &+\epsilon_S(h\circ w^*;h^*\circ w^*)\\
    &+\sup_{h,h^*\in\mathcal{H}}|\epsilon_T(h\circ w^*;h^*\circ w^*)-\epsilon_S(h\circ w^*;h^*\circ w^*)|\\
    \leq& \epsilon_T(h^*\circ w^*;f_T)+\epsilon_T(h\circ w;h\circ w^*)\\
    &+\epsilon_S(h\circ w^*;h^*\circ w^*)+\frac{1}{2}d_{\mathcal{H}\Delta\mathcal{H}}(\mathcal{D}_S^Z,\mathcal{D}_T^Z)\\
    \leq& \epsilon_T(h^*\circ w^*;f_T)+\epsilon_S(h\circ w^*;h^*\circ w^*)\\
    &+\epsilon_S(h\circ w;h\circ w^*)+\frac{1}{2}d_{\mathcal{H}\Delta\mathcal{H}}(\mathcal{D}_S^Z,\mathcal{D}_T^Z)\\
    &+|\epsilon_T(h\circ w;h\circ w^*)-\epsilon_S(h\circ w;h\circ w^*)|\\
    \leq& \epsilon_T(h^*\circ w^*;f_T)+\epsilon_S(h^*\circ w^*;f_S)+\epsilon_S(h\circ w;f_S)\\&+2\epsilon_S(h\circ w^*,f_S)+\frac{1}{2}d_{\mathcal{H}\Delta\mathcal{H}}(\mathcal{D}_S^Z,\mathcal{D}_T^Z)\\
    &+\sup_{w,w^*\in\mathcal{W}}|\epsilon_T(h\circ w;h\circ w^*)-\epsilon_S(h\circ w;h\circ w^*)|\\
    \leq& \epsilon_T(h^*\circ w^*;f_T)+\epsilon_S(h^*\circ w^*;f_S)+\epsilon_S(h\circ w;f_S)\\&+2\epsilon_S(h\circ w^*,f_S)+\frac{1}{2}d_{\mathcal{H}\Delta\mathcal{H}}(\mathcal{D}_S^Z,\mathcal{D}_T^Z)\\
    &+\sup_{h\in\mathcal{H}}\big[\frac{1}{2}d_{h,\mathcal{W}\Delta\mathcal{W}}(\mathcal{D}_S^X,\mathcal{D}_T^X)\big]\\
    \leq& \epsilon_S(h\circ w;f_S)+2\epsilon_S(h\circ w^*,f_S)\\
    &+\frac{1}{2}d_{\mathcal{H}\Delta\mathcal{H}}({Z}_T,{Z}_S)\\
    &+4\sqrt{\frac{2d_H\log{(2N)+\log(2/\delta)}}{N}}\\
    &+\frac{1}{2}\sup_{\hat{h}\in\mathcal{H}}\big[d_{\hat{h},\mathcal{W}\Delta\mathcal{W}}({X}_S,{X}_T)\big]\\
    &+4\sqrt{\frac{2d_W\log{(2N)+\log(2/\delta)}}{N}}\\
    &+\epsilon_T(h^*\circ w^*;f_T)+\epsilon_S(h^*\circ w^*;f_S).
    \end{align*}
\endgroup
\end{proof}
\end{theorem}

\end{document}